\documentclass[journal]{IEEEtran}
\usepackage{cite}
\usepackage{scrextend}
\usepackage{hyperref}
\usepackage{tikz}
\usepackage{tabularx}
\usepackage{booktabs}
\usepackage{makecell}
\usepackage{ifthen}
\usepackage[official]{eurosym}

\usetikzlibrary{shapes}
\usetikzlibrary{3d}

\newcommand{\networkLayer}[6]{\def\a{#1}
\def\b{0.02}
\def\c{#2}
\def\t{#3}
\def\d{#4}
\draw[line width=0.3mm] (\c+\t,0,\d) -- (\c+\t,\a,\d) -- (\t,\a,\d);                                                      
\draw[line width=0.3mm] (\t,0,\a+\d) -- (\c+\t,0,\a+\d) node[midway,below] {#6} -- (\c+\t,\a,\a+\d) -- (\t,\a,\a+\d) -- (\t,0,\a+\d); 
\draw[line width=0.3mm] (\c+\t,0,\d) -- (\c+\t,0,\a+\d);
\draw[line width=0.3mm] (\c+\t,\a,\d) -- (\c+\t,\a,\a+\d);
\draw[line width=0.3mm] (\t,\a,\d) -- (\t,\a,\a+\d);
\filldraw[#5] (\t+\b,\b,\a+\d) -- (\c+\t-\b,\b,\a+\d) -- (\c+\t-\b,\a-\b,\a+\d) -- (\t+\b,\a-\b,\a+\d) -- (\t+\b,\b,\a+\d); 
\filldraw[#5] (\t+\b,\a,\a-\b+\d) -- (\c+\t-\b,\a,\a-\b+\d) -- (\c+\t-\b,\a,\b+\d) -- (\t+\b,\a,\b+\d);
\ifthenelse {\equal{#5} {}}
{}
{\filldraw[#5] (\c+\t,\b,\a-\b+\d) -- (\c+\t,\b,\b+\d) -- (\c+\t,\a-\b,\b+\d) -- (\c+\t,\a-\b,\a-\b+\d);} 
}

\begin{document}
\title{Deep Learning in Cardiology}
\author{Paschalis~Bizopoulos, and~Dimitrios~Koutsouris
\thanks{This work was supported by the European Union's Horizon 2020 research and innovation programme under Grant agreement 727521. 

The authors are with Biomedical Engineering Laboratory, School of Electrical and Computer Engineering, National Technical University of Athens, Athens 15780, Greece e-mail: pbizop@gmail.com, dkoutsou@biomed.ntua.gr, (Corresponding author: Paschalis Bizopoulos).}}

\IEEEoverridecommandlockouts
\IEEEpubid{\makebox[\columnwidth]{10.1109/RBME.2018.2885714/\$33.00~\ copyright~\copyright~2019~IEEE \hfill} \hspace{\columnsep}\makebox[\columnwidth]{ }}
\maketitle
\IEEEpubidadjcol

\begin{abstract}
	The medical field is creating large amount of data that physicians are unable to decipher and use efficiently.
	Moreover, rule-based expert systems are inefficient in solving complicated medical tasks or for creating insights using big data.
	Deep learning has emerged as a more accurate and effective technology in a wide range of medical problems such as diagnosis, prediction and intervention.
	Deep learning is a representation learning method that consists of layers that transform the data non-linearly, thus, revealing hierarchical relationships and structures.
	In this review we survey deep learning application papers that use structured data, signal and imaging modalities from cardiology.
	We discuss the advantages and limitations of applying deep learning in cardiology that also apply in medicine in general, while proposing certain directions as the most viable for clinical use.
\end{abstract}

\begin{IEEEkeywords}
	Deep learning, neural networks, cardiology, cardiovascular disease
\end{IEEEkeywords}

\section{Introduction}
\label{sec:introduction}
\IEEEPARstart{C}{ardiovascular} Diseases (CVDs), are the leading cause of death worldwide accounting for 30\% of deaths in 2014 in United States\cite{benjamin2017heart}, 45\% of deaths in Europe and they are estimated to cost \euro{210} billion per year just for the European Union\cite{wilkins2017european}.
Physicians are still taught to diagnose CVDs based on medical history, biomarkers, simple scores, and physical examinations of individual patients which they interpret according to their personal clinical experience.
Then, they match each patient to the traditional taxonomy of medical conditions according to a subjective interpretation of the medical literature.
This kind of procedure is increasingly proven error-prone and inefficient.
Moreover, cardiovascular technologies constantly increase their capacity to capture large quantities of data, making the work of physicians more demanding.
Therefore, automating medical procedures is required for increasing the quality of health of patients and decreasing the cost of healthcare systems.

The need for automating medical procedures ranges from diagnosis to treatment and in cases where there is lack of healthcare service from physicians.
Previous efforts include rule-based expert systems, designed to imitate the procedure that medical experts follow when solving medical tasks or creating insights.
These systems are proven to be inefficient because they require significant feature engineering and domain knowledge to achieve adequate accuracy and they are hard to scale in presence of new data.

Machine learning is a set of artificial intelligence (AI) methods that allows computers learn a task using data, instead of being explicitly programmed.
It has emerged as an effective way of using and combining biomarkers, imaging, aggregated clinical research from literature, and physician's notes from Electronic Health Records (EHRs) to increase the accuracy of a wide range of medical tasks.
Medical procedures using machine learning are evolving from art to data-driven science, bringing insight from population-level data to the medical condition of the individual patient.

Deep learning, and its application on neural networks Deep Neural Networks (DNNs), is a set of machine learning methods that consist of multiple stacked layers and use data to capture hierarchical levels of abstraction.
Deep learning has emerged due to the increase of computational power of Graphical Processing Units (GPUs) and the availability of big data and has proven to be a robust solution for tasks such as image classification\cite{krizhevsky2012imagenet}, image segmentation\cite{ronneberger2015u}, natural language processing\cite{collobert2008unified}, speech recognition\cite{graves2013speech} and genomics\cite{alipanahi2015predicting}.

Advantages of DNNs against traditional machine learning techniques include that they require less domain knowledge for the problem they are trying to solve and they are easier to scale because increase in accuracy is usually achieved either by increasing the training dataset or the capacity of the network.
Shallow learning models such as decision trees and Support Vector Machines (SVMs) are `inefficient'; meaning that they require a large number of computations during training/inference, large number of observations for achieving generalizability and significant human labour to specify prior knowledge in the model\cite{bengio2007scaling}.

In this review we present deep learning applications in structured data, signal and imaging modalities from cardiology, related to heart and vessel structures.
The literature phrase search is the combined presence of each one of the cardiology terms indicated by ($*$) in Table~\ref{table:cardiologyacronyms} with each one of the deep learning terms related to architecture, indicated by ($+$) in Table~\ref{table:deeplearningacronyms} using Google Scholar\footnote{\url{https://scholar.google.com}}, Pubmed\footnote{\url{https://ncbi.nlm.nih.gov/pubmed/}} and Scopus\footnote{\url{https://www.scopus.com/search/form.uri?=display=basic}}.
Results are then curated to match the selection criteria of the review and summarized according to two main axis: neural network architecture and the type of data that was used for training/validation/testing.
Evaluations are reported for areas that used a consistent set of metrics with the same unaltered database with the same research question.
Papers that do not provide information on the neural network architecture or papers that duplicate methods of previous work or preliminary papers are excluded from the review.
When \textit{multiple} is reported in the Results column in Tables~\ref{table:structured},~\ref{table:signals1},~\ref{table:signals2},~\ref{table:imaging1},~\ref{table:imaging2},~\ref{table:imaging3}, they are reported in the main text where suitable.
Additionally, for CNN architectures the use of term \textit{layer} implies `convolutional layer' for the sake of brevity.

At Section~\ref{sec:neuralnetworks}, we present the fundamental concepts of neural networks and deep learning along with an overview of the architectures that have been used in cardiology.
Then at Sections~\ref{sec:structured},~\ref{sec:signals} and~\ref{sec:imaging} we present the deep learning applications using structured data, signal and imaging modalities from cardiology respectively.
Table~\ref{table:cardiologypublicdatabases} provides an overview of the publicly available cardiology databases that have been used with deep learning.
Finally at Section~\ref{sec:discussion} we present the specific advantages and limitations of deep learning applications in cardiology and we conclude proposing certain directions for making deep learning applicable to clinical use.

\begin{table}[!t]
	\caption{Cardiology acronyms}
	\label{table:cardiologyacronyms}
	\begin{minipage}{0.5\textwidth}
		\centering
		\begin{tabularx}{\textwidth}{c l}
			\toprule
			\thead{Acronym\footnote{($*$) denotes the term was used in the literature phrase search in combination with those from Table~\ref{table:deeplearningacronyms}.}} & \thead{Meaning}                                            \\
			\midrule
			ACDC                                                                                                                                                             & Automated Cardiac Diagnosis Challenge                      \\
			ACS$^*$                                                                                                                                                          & Acute Coronary Syndrome                                    \\
			AF$^*$                                                                                                                                                           & Atrial Fibrillation                                        \\
			BIH                                                                                                                                                              & Beth Israel Hospital                                       \\
			BP                                                                                                                                                               & Blood Pressure                                             \\
			CAC                                                                                                                                                              & Coronary Artery Calcification                              \\
			CAD$^*$                                                                                                                                                          & Coronary Artery Disease                                    \\
			CHF$^*$                                                                                                                                                          & Congestive Heart Failure                                   \\
			CT$^*$                                                                                                                                                           & Computerized Tomography                                    \\
			CVD$^*$                                                                                                                                                          & Cardiovascular Disease                                     \\
			DBP                                                                                                                                                              & Diastolic Blood Pressure                                   \\
			DWI$^*$                                                                                                                                                          & Diffusion Weighted Imaging                                 \\
			ECG$^*$                                                                                                                                                          & Electrocardiogram                                          \\
			EHR$^*$                                                                                                                                                          & Electronic Health Record                                   \\
			FECG$^*$                                                                                                                                                         & Fetal ECG                                                  \\
			HF$^*$                                                                                                                                                           & Heart Failure                                              \\
			HT                                                                                                                                                               & Hemorrhagic Transformation                                 \\
			HVSMR                                                                                                                                                            & Heart \& Vessel Segmentation from 3D MRI                   \\
			ICD                                                                                                                                                              & International Classification of Diseases                   \\
			IVUS$^*$                                                                                                                                                         & Intravascular Ultrasound                                   \\
			LV                                                                                                                                                               & Left Ventricle                                             \\
			MA                                                                                                                                                               & Microaneurysm                                              \\
			MI$^*$                                                                                                                                                           & Myocardial Infarction                                      \\
			MMWHS                                                                                                                                                            & Multi-Modality Whole Heart Segmentation Challenge          \\
			MRA$^*$                                                                                                                                                          & Magnetic Resonance Angiography                             \\
			MRI$^*$                                                                                                                                                          & Magnetic Resonance Imaging                                 \\
			MRP$^*$                                                                                                                                                          & Magnetic Resonance Perfusion                               \\
			OCT$^*$                                                                                                                                                          & Optical Coherence Tomography                               \\
			PCG$^*$                                                                                                                                                          & Phonocardiogram                                            \\
			PPG$^*$                                                                                                                                                          & Pulsatile Photoplethysmography                             \\
			RV                                                                                                                                                               & Right Ventricle                                            \\
			SBP                                                                                                                                                              & Systolic Blood Pressure                                    \\
			SLO$^*$                                                                                                                                                          & Scanning Laser Ophthalmoscopy                               \\
			STACOM                                                                                                                                                           & Statistical Atlases \& Computational Modeling of the Heart \\
			\bottomrule
		\end{tabularx}
	\end{minipage}
\end{table}

\begin{table}[!t]
	\caption{Deep learning acronyms}
	\label{table:deeplearningacronyms}
	\begin{minipage}{0.5\textwidth}
		\centering
		\begin{tabularx}{\textwidth}{c l}
			\toprule
			\thead{Acronym\footnote{($+$) denotes the term was used in the literature phrase search in combination with those from Table~\ref{table:cardiologyacronyms}.}} & \thead{Meaning}                                           \\
			\midrule
			AE$^+$                                                                                                                                                         & Autoencoder                                               \\
			AUC                                                                                                                                                            & Area Under Curve                                          \\
			AI$^+$                                                                                                                                                         & Artificial Intelligence                                   \\
			CNN$^+$                                                                                                                                                        & Convolutional Neural Network                              \\
			CRF                                                                                                                                                            & Conditional Random Field                                  \\
			DBN$^+$                                                                                                                                                        & Deep Belief Network                                       \\
			DNN$^+$                                                                                                                                                        & Deep Neural Network                                       \\
			FCN$^+$                                                                                                                                                        & Fully Convolutional Network                               \\
			FNN$^+$                                                                                                                                                        & Fully Connected Network                                   \\
			GAN$^+$                                                                                                                                                        & Generative Adversarial Network                            \\
			GRU$^+$                                                                                                                                                        & Gated Recurrent Unit                                      \\
			LSTM$^+$                                                                                                                                                       & Long-Short Term Memory                                    \\
			MFCC                                                                                                                                                           & Mel-Frequency Cepstral Coefficient                        \\
			MICCAI                                                                                                                                                         & Medical Image Computing \& Computer-Assisted Intervention \\
			PCA                                                                                                                                                            & Principal Component Analysis                              \\
			RBM$^+$                                                                                                                                                        & Restricted Boltzmann Machine                              \\
			RF                                                                                                                                                             & Random Forest                                             \\
			RNN$^+$                                                                                                                                                        & Recurrent Neural Network                                  \\
			ROI                                                                                                                                                            & Region of Interest                                        \\
			SAE$^+$                                                                                                                                                        & Stacked Autoencoder                                       \\
			SATA                                                                                                                                                           & Segmentation Algorithms, Theory and Applications          \\
			SDAE$^+$                                                                                                                                                       & Stacked Denoised Autoencoder                              \\
			SSAE$^+$                                                                                                                                                       & Stacked Sparse Autoencoder                                \\
			SVM                                                                                                                                                            & Support Vector Machine                                    \\
			VGG                                                                                                                                                            & Visual Geometry Group                                     \\
			WT                                                                                                                                                             & Wavelet Transform                                         \\
			\bottomrule
		\end{tabularx}
	\end{minipage}
\end{table}

\section{Neural networks}
\label{sec:neuralnetworks}
\subsection{Theory overview}
Neural networks is a set of machine learning techniques initially inspired by the brain but without a primary aim to simulate it.
They are function approximation methods where the input $\mathbf{x}$ is text, image, sound, generic signal, 3D volume, video (or a combination of these) and the output $\mathbf{y}$ is from the same set as $\mathbf{x}$ but with a more informative content.
In mathematical terms the objective of a neural network is to find the set of parameters $\theta$ (weights $\mathbf{w}$ and biases $\mathbf{b}$):
\begin{equation}
	\centering
	f(\mathbf{x};\mathbf{\theta}) = \mathbf{\hat{y}}
\end{equation}

\noindent
where $f$ is a predefined function and $\mathbf{\hat{y}}$ is the prediction.
The constraint for $\theta$ is to have an as low as possible result for a cost function $J(\theta)$ between $\mathbf{y}$ and $\mathbf{\hat{y}}$.

\begin{figure}[!t]
	\centering
	\begin{tikzpicture}[]
		\node[circle, draw=white] (x1) at (0, 4) {$x_1$};
		\node at (0, 3) {$\vdots$};
		\node[circle, draw=white] (xj) at (0, 2) {$x_j$};
		\node at (0, 1) {$\vdots$};
		\node[circle, draw=white] (xn) at (0, 0) {$x_n$};
		\node[circle, draw=white] (b) at (4, 4) {$b$};
		\node[circle, draw=black] (sigma) at (4, 2) {$\phi$};
		\node[circle, draw=white] (output) at (6, 2) {$\alpha$};
		\draw[->] (x1) to node[above]{$w_1$} (sigma);
		\draw[->] (xj) to node[above]{$w_j$} (sigma);
		\draw[->] (xn) to node[above]{$w_n$} (sigma);
		\draw[->] (b) -- (sigma);
		\draw[->] (sigma) -- (output);
	\end{tikzpicture}
	\caption{Perceptron.
	It consists of a set of connections $x_{1, \ldots, n}$ as its input, the weights $w_{1, \ldots, n}$, the bias $b$, the activation function $\phi$ and the output $\alpha$.
	}
	\label{fig:perceptron}
\end{figure}
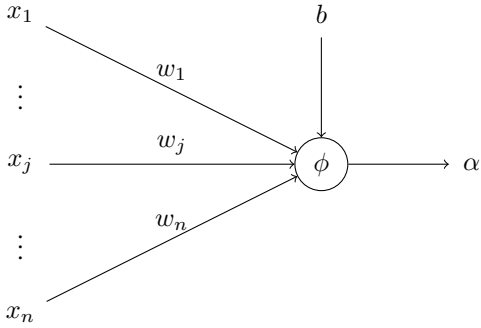

The basic unit of neural networks is the perceptron depicted in Fig.\ref{fig:perceptron} which was first published by Rosenblatt\cite{rosenblatt1958perceptron} in 1958.
It consists of a set of connections with the vector $\mathbf{x}=[x_1, \ldots, x_j, \ldots, x_n]$ as its input, while the strength of the connections, also called weights, $\mathbf{w}=[w_1, \ldots, w_j, \ldots, w_n]$ and the bias $b$, are learned iteratively from $\mathbf{x}$ and $\mathbf{y}$.
The node's decision to fire a signal $\alpha$ to the next neuron or to the output is determined by the activation function $\phi$, the weighted sum of $\mathbf{w}$ and $\mathbf{x}$ and the bias $b$, in the following way:
\begin{equation}
	\centering
	\alpha = \phi(\sum\limits_{j} w_{j}x_{j} + b)
\end{equation}

The weights $\mathbf{w}$ and biases $\mathbf{b}$ in DNNs are adjusted iteratively using gradient descent based optimization algorithms and backpropagation\cite{rumelhart1986learning} which calculates the gradient of the cost function with respect to the parameters $\nabla_{\theta}J(\theta)$\cite{goodfellow2016deep}.
Evaluating the generalization of a neural network requires splitting the initial dataset $D=(\mathbf{x}, \mathbf{y})$ into three non-overlapping datasets:
\begin{equation}
	\centering
	D_{train}=(\mathbf{x}_{train}, \mathbf{y}_{train})
\end{equation}

\begin{equation}
	\centering
	D_{validation}=(\mathbf{x}_{validation}, \mathbf{y}_{validation})
\end{equation}

\begin{equation}
	\centering
	D_{test}=(\mathbf{x}_{test}, \mathbf{y}_{test})
\end{equation}
$D_{train}$ is used for adjusting $\mathbf{w}$ and $\mathbf{b}$ that minimize the chosen cost function while $D_{validation}$ is used for choosing the hyperparameters of the network.
$D_{test}$ is used to evaluate the generalization of the network and it should ideally originate from different machines/patients/organizations depending on the research question that is targeted.

The performance of DNNs has improved using the Rectified Linear Unit (ReLU) as an activation function compared to the logistic sigmoid and hyperbolic tangent\cite{glorot2011deep}.
The activation function of the last layer depends on the nature of research question to be answered (e.g.\ softmax for classification, sigmoid for regression).

The cost functions that are used in neural networks also depend on the task to be solved.
Cross entropy quantifies the difference between the true and predicted probability distributions and is usually chosen for detection and classification problems.
The Area Under the receiver operating Curve (AUC) represents the probability that a random pair of normal and abnormal pixels/signals/images will be correctly classified\cite{hanley1982meaning} and is used in binary segmentation problems.
Dice coefficient\cite{dice1945measures} is a measure of similarity used in segmentation problems and its values range between zero (total mismatch) to unity (perfect match).

\subsection{Architectures overview}
Fully Connected Networks (FNNs) are networks that consist of multiple perceptrons stacked in width and depth, meaning that every unit in each layer is connected to every unit in the layers immediately before and after.
Although it has been proven\cite{hornik1989multilayer} that one layer FNNs with sufficient number of hidden units are universal function approximators, they are not computationally efficient for fitting complex functions.
Deep Belief Networks (DBNs)\cite{hinton2006fast} are stacked Restricted Boltzmann Machines (RBMs) where each layer encodes statistical dependencies among the units in the previous layer; they are trained to maximize the likelihood of the training data.

\begin{figure*}[!t]
	\centering
	\begin{tikzpicture}[scale=1.1]
		\draw[dashed] (0.03, 0) -- (1.5, 0);
		\draw[dashed] (1.61, 0) -- (3.01, 0);
		\draw[dashed] (3.2, 0) -- (4.5, 0);
		\draw[dashed] (4.9, 0) -- (6, 0);
		\draw[dashed] (6.8, 0) -- (7.5, 0);
		\networkLayer{4.0}{0.03}{0}{0.0}{color=white}{}
		\networkLayer{3.5}{0.1}{1.5}{0.0}{color=white}{}
		\networkLayer{3.0}{0.2}{3.0}{0.0}{color=white}{}
		\networkLayer{2.5}{0.4}{4.5}{0.0}{color=white}{}
		\networkLayer{2.0}{0.8}{6.0}{0.0}{color=white}{}
		\networkLayer{1.5}{1.6}{7.5}{0.0}{color=white}{}
		\draw[dashed] (10, -1) -- (11.5, 0.5);
		\draw[-] (10.5, -1) -- (12, 0.5);
		\draw[-] (10.5, -0.5) -- (12, 1);
		\draw[-] (10, -0.5) -- (11.5, 1);
		\draw[-] (12, 1) -- (11.5, 1);
		\draw[-] (12, 1) -- (12, 0.5);
		\draw[-] (10, -1) rectangle ++ (0.5,0.5) node{};
		\node[canvas is zy plane at x=0, scale=1.1] at (-0.735, 1.235){\includegraphics[scale=0.491]{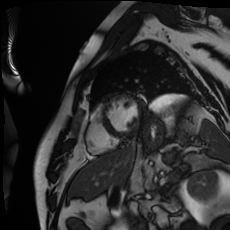}};
		\node[canvas is zy plane at x=0, scale=1.1] at (0.93, 1.07){\includegraphics[scale=0.89]{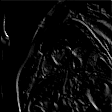}};
		\node[canvas is zy plane at x=0, scale=1.1, opacity=0.5] at (0.83, 1.07){\includegraphics[scale=0.89]{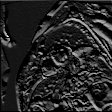}};
		\node[canvas is zy plane at x=0, scale=1.1] at (2.63, 0.93){\includegraphics[scale=1.52]{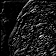}};
		\node[canvas is zy plane at x=0, scale=1.1, opacity=0.5] at (2.42, 0.93){\includegraphics[scale=1.52]{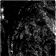}};
		\node[canvas is zy plane at x=0, scale=1.1] at (4.42, 0.78){\includegraphics[scale=2.53]{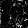}};
		\node[canvas is zy plane at x=0, scale=1.1, opacity=0.5] at (4.03, 0.78){\includegraphics[scale=2.53]{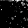}};
		\node[canvas is zy plane at x=0, scale=1.1] at (6.42, 0.611){\includegraphics[scale=4.05]{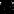}};
		\node[canvas is zy plane at x=0, scale=1.1, opacity=0.5] at (5.6, 0.611){\includegraphics[scale=4.05]{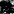}};
		\node[canvas is zy plane at x=0, scale=1.1] at (8.815, 0.472){\includegraphics[scale=6.05]{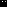}};
		\node[canvas is zy plane at x=0, scale=1.1, opacity=0.5] at (7.22, 0.472){\includegraphics[scale=6.05]{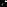}};
		\node at (8, 0.472) {$\dots$};
		\node at (6, 0.472) {$\dots$};
		\node[canvas is zy plane at x=0] at (-1, -3){\includegraphics[scale=10]{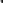}};
		\node at (-0.8, -3) {$\dots$};
		\node[canvas is zy plane at x=0] at (-0.6, -3){\includegraphics[scale=10]{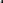}};
		\node[canvas is zy plane at x=0] at (0.8, -3){\includegraphics[scale=10]{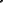}};
		\node at (1, -3) {$\dots$};
		\node[canvas is zy plane at x=0] at (1.3, -3){\includegraphics[scale=10]{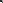}};
		\node[canvas is zy plane at x=0] at (2.5, -3){\includegraphics[scale=10]{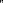}};
		\node at (2.7, -3) {$\dots$};
		\node[canvas is zy plane at x=0] at (2.9, -3){\includegraphics[scale=10]{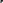}};
		\node[canvas is zy plane at x=0] at (4.2, -3){\includegraphics[scale=10]{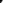}};
		\node at (4.4, -3) {$\dots$};
		\node[canvas is zy plane at x=0] at (4.6, -3){\includegraphics[scale=10]{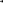}};
		\node[canvas is zy plane at x=0] at (6.3, -3){\includegraphics[scale=10]{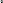}};
		\node at (6.5, -3) {$\dots$};
		\node[canvas is zy plane at x=0] at (6.7, -3){\includegraphics[scale=10]{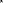}};
		\draw[<-] (-0.3, -3) -- (0.3, -3);
		\draw[<-] (1.6, -3) -- (2.1, -3);
		\draw[<-] (3.2, -3) -- (3.9, -3);
		\draw[<-] (4.8, -3) -- (5.9, -3);
		\draw[<-] (7.1, -3) -- (13.0, -3);
		\draw[dashed] (-1.51, -1.53) -- (0.15, -1.35);
		\draw[dashed] (0.27, -1.35) -- (1.85, -1.16);
		\draw[dashed] (2.05, -1.16) -- (3.55, -0.97);
		\draw[dashed] (3.95, -0.97) -- (5.25, -0.77);
		\draw[dashed] (6.02, -0.78) -- (6.94, -0.57);
		\draw[dashed] (-1.51, 2.45) -- (0.15, 2.15);
		\draw[dashed] (0.27, 2.15) -- (1.85, 1.85);
		\draw[dashed] (2.05, 1.85) -- (3.55, 1.54);
		\draw[dashed] (3.95, 1.54) -- (5.25, 1.23);
		\draw[dashed] (6.02, 1.22) -- (6.94, 0.91);
		\draw[dashed] (0.03, 4) -- (1.5, 3.51);
		\draw[dashed] (1.61, 3.51) -- (3.01, 3.01);
		\draw[dashed] (3.2, 3) -- (4.5, 2.5);
		\draw[dashed] (4.9, 2.51) -- (6, 2);
		\draw[dashed] (6.8, 2) -- (7.5, 1.5);
		\draw[dotted] (9.1, 1.5) --(11.5, 1);
		\draw[dotted] (9.1, 0) -- (11.5, 0.5);
		\draw[dotted] (8.5, 0.92) -- (10, -0.5);
		\draw[dotted] (8.5, -0.58) --  (10, -1);
		\draw[dashdotted] (10.5, -0.5) -- (13, -0.25);
		\draw[dashdotted] (10.5, -1) -- (13, -0.25);
		\draw[dashdotted] (12, 1) -- (13, -0.25);
		\draw[dashdotted] (12, 0.5) -- (13, -0.25);
		\node[align=left] at (13.3,-0.2) {$\hat{y}$};
		\draw[->] (13.3, -0.6) -- (13.3, -2.7) node [pos=1,below=1] {$J$};
		\draw[<-] (13.5, -3) -- (14, -3) node [pos=1,right=1] {$y$};
		\node[align=left] at (10.5, -2.4) {Backpropagation};
		\node[align=left] at (10.8, 2.5) {Feed-forward};
	\end{tikzpicture}
	\caption{A Convolutional Neural Network that calculates the LV area ($\hat{y}$) from an MRI image ($x$).
	The pyramidoid structure on top denotes the flow of the feed-forward calculations starting from input image $x$ through the set of feature maps depicted as 3D rectangulars to the output $\hat{y}$.
	The height and width of the set of feature maps is proportional to the height and width of the feature maps while the depth is proportional to the number of the feature maps.
	The arrows at the bottom denote the flow of the backpropagation starting after the calculation of the loss using the cost function $J$, the original output $y$ and the predicted output $\hat{y}$.
	This loss is backpropagated through the filters of the network adjusting their weight.
	Dashed lines denote a 2D convolutional layer with ReLU and Max-Pooling (which also reduces the height and width of the feature maps), the dotted line denotes the fully connected layer and the dash dotted lines at the end denote the sigmoid layer.
	For visualization purposes only a few of the feature maps and filters are shown, and they are also not in scale.}
	\label{fig:cnn}
\end{figure*}
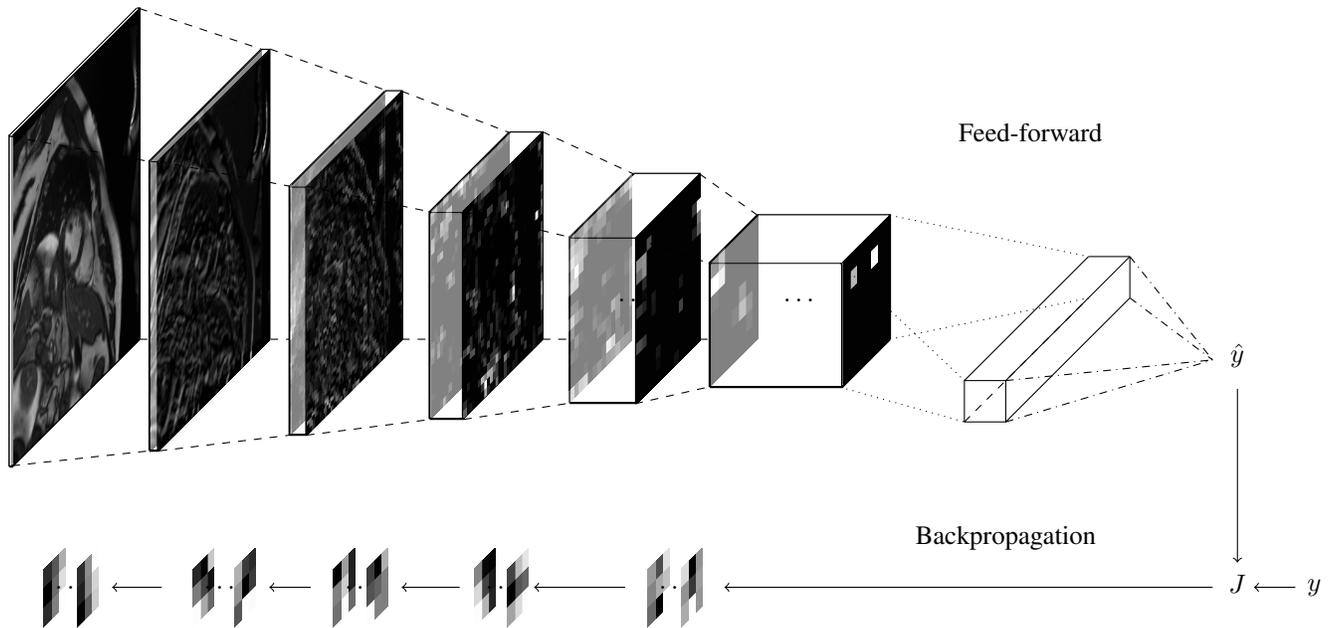

Convolutional Neural Networks (CNNs), as shown in Fig.~\ref{fig:cnn}, consist of a convolutional part where hierarchical feature extraction takes place (low-level features such as edges and corners and high-level features such as parts of objects) and a fully connected part for classification or regression, depending on the nature of the output $y$.
Convolutional layers are much better feature optimizers that utilize the local relationships in the data, while fully connected layers are good classifiers, thus they are used as the last layers of a CNN\@.
Additionally, convolutional layers create feature maps using shared weights that have a fixed number of parameters in contrast with fully connected layers, making them much faster.
VGG\cite{simonyan2014very} is a simple CNN architecture that utilizes small convolutional filters ($3\times 3$) and performance is increased by increasing the depth of the network.
GoogleNet\cite{szegedy2015going} is another CNN-like architecture that makes use of the inception module.
The inception module uses multiple convolutional layers in parallel from which the result is the concatenated, thus allowing the network to learn multiple level features.
ResNet\cite{he2016deep} is a CNN-like architecture that formulates layers as learning residual functions with reference to the layer inputs, allowing training of much deeper networks.

Autoencoders (AEs) are neural networks that are trained with the objective to copy the input $x$ to the output in such a way that they encode useful properties of the data.
It usually consists of an encoding part that downsamples the input down to a linear feature and a decoding part that upsamples to the original dimensions.
A common AE architecture is Stacked Denoised AE (SDAE) that has an objective to reconstruct the clean input from an artificially corrupted version of the input\cite{vincent2010stacked} which prevents the model from learning trivial solutions.
Another AE-like architecture is u-net\cite{ronneberger2015u}, which is of special interest to the biomedical community since it was first applied on segmentation of biomedical images.
U-net introduced skip connections that connect the layers of the encoder with corresponding ones from the decoder.

Recurrent Neural Networks (RNNs) are networks that consist of feedback loops and in contrast to previously defined architectures they can use their internal state to process the input.
Vanilla RNNs have the vanishing gradients problem and for that reason Long-Short Term Memory (LSTM)\cite{hochreiter1997long} was proposed as a solution to storing information over extended time.
Gated Recurrent Unit (GRU)\cite{cho2014properties} was later proposed as a simpler alternative to LSTM\@.

\begin{table*}[!t]
	\caption{Cardiology Public Databases}
	\label{table:cardiologypublicdatabases}
	\begin{minipage}{\textwidth}
		\centering
		\begin{tabularx}{\textwidth}{l c r l}
			\toprule
			\thead{Database Name\footnote{URLs for each database are provided in the reference section.}} & \thead{Acronym} & \thead{Patients}                                                                  & \thead{Task}                                                 \\
			\midrule
			\multicolumn{4}{l}{\thead{Structured Databases}}                                                                                                                                                                                                                   \\
			\midrule
			Medical Information Mart for Intensive Care III\cite{johnson2016mimic}                        & MIMIC           & 38597                                                                             & 53423 hospital admissions for ICD-9 and mortality prediction \\
			KNHANES-VI\cite{kweon2014data}                                                                & KNH             & 8108                                                                              & epidemiology tasks with demographics, blood tests, lifestyle \\
			\midrule
			\multicolumn{4}{l}{\thead{Signal Databases (all ECG besides\cite{karlen2013multiparameter})}}                                                                                                                                                                      \\
			\midrule
			IEEE-TBME PPG Respiratory Rate Benchmark Dataset\cite{karlen2013multiparameter}               & PPGDB           & 42                                                                                & respiratory rate estimation using PPG                        \\
			Creighton University Ventricular Tachyarrhythmia\cite{nolle1986crei}                          & CREI            & 35                                                                                & ventricular tachyarrhythmia detection                        \\
			MIT-BIH Atrial Fibrillation Database\cite{moody1983new}                                        & AFDB            & 25                                                                                & AF prediction                                                \\
			BIH Deaconess Medical Center CHF Database\cite{baim1986survival}                              & CHFDB           & 15                                                                                & CHF classification                                           \\
			St.Petersburg Institute of Cardiological Technics\cite{goldberger2000physiobank}               & INDB            & 32                                                                                & QRS detection and ECG beat classification                    \\
			Long-Term Atrial Fibrillation Database\cite{petrutiu2007abrupt}                                & LTAFDB          & 84                                                                                & QRS detection and ECG beat classification                    \\
			Long-Term ST Database\cite{jager2003long}                                                     & LTSTDB          & 80                                                                                & ST beat detection and classification                         \\
			MIT-BIH Arrhythmia Database\cite{moody2001impact}                                             & MITDB           & 47                                                                                & arrhythmia detection                                         \\
			MIT-BIH Noise Stress Test Database\cite{moody1984noise}                                       & NSTDB           & 12                                                                                & used for noise resilience tests of models                    \\
			MIT-BIH Normal Sinus Rhythm Database\cite{goldberger2000physiobank}                           & NSRDB           & 18                                                                                & arrhythmia detection                                         \\
			MIT-BIH Normal Sinus Rhythm RR Interval Database\cite{goldsmith1992comparison}                & NSR2DB          & 54                                                                                & ECG beat classification                                      \\
			Fantasia Database\cite{iyengar1996age}                                                        & FAN             & 40                                                                                & ECG beat classification                                      \\
			AF Classification short single lead ECG Physionet 2017\cite{moody2001impact}                  & PHY17           & ---\footnote{\label{publicdatabaselabel}The number of patients was not reported.} & ECG beat classification (12186 single lead records)          \\
			Physionet 2016 Challenge\cite{liu2016open}                                                    & PHY16           & 1297                                                                              & heart sound classification using PCG (3126 records)          \\
			Physikalisch-Technische Bundesanstalt ECG Database\cite{bousseljot1995nutzung}                & PTBDB           & 268                                                                               & cardiovascular disease diagnosis                             \\
			QT Database\cite{laguna1997database}                                                          & QTDB            & 105                                                                               & QT beat detection and classification                         \\
			MIT-BIH Supraventricular Arrhythmia Database\cite{greenwald1990improved}                      & SVDB            & 78                                                                                & supraventricular arrhythmia detection                        \\
			Non-Invasive Fetal ECG Physionet Challenge Dataset\cite{silva2013noninvasive}                 & PHY13           & 447                                                                               & measurement of fetal HR, RR interval and QT                  \\
			DeepQ Arrhythmia Database\cite{wu2017deepq}\footnote{Authors mention that they plan to make the database publicly available.}                 & DeepQ           & 299                                                                               & ECG beat classification(897 records)                 \\
			\midrule
			\multicolumn{4}{l}{\thead{MRI Databases}}                                                                                                                                                                                                                          \\
			\midrule
			MICCAI 2009 Sunnybrook\cite{radau2009evaluation}                                              & SUN09           & 45                                                                                & LV segmentation                                              \\
			MICCAI 2011 Left Ventricle Segmentation STACOM\cite{fonseca2011cardiac}                       & STA11           & 200                                                                               & LV segmentation                                              \\
			MICCAI 2012 Right Ventricle Segmentation Challenge\cite{petitjean2015right}                   & RV12            & 48                                                                                & RV segmentation                                              \\
			MICCAI 2013 SATA\cite{asman2013miccai}                                                        & SAT13           & ---\footref{publicdatabaselabel}                                                  & LV segmentation                                              \\
			MICCAI 2016 HVSMR\cite{pace2015interactive}                                                   & HVS16           & 20                                                                                & whole heart segmentation                                     \\
			MICCAI 2017 ACDC\cite{bernard2018deep}                                                        & AC17            & 150                                                                               & LV/RV segmentation                                            \\
			York University Database\cite{andreopoulos2008efficient}                                      & YUDB            & 33                                                                                & LV segmentation                                              \\
			Data Science Bowl Cardiac Challenge Data\cite{dsbcdc2016}                                     & DS16            & 1140                                                                              & LV volume estimation after systole and diastole              \\
			\midrule
			\multicolumn{4}{l}{\thead{Retina Databases (all Fundus besides\cite{zhang2016robust})}}                                                                                                                                                                            \\
			\midrule
			Digital Retinal Images for Vessel Extraction\cite{staal2004ridge}                             & DRIVE           & 40                                                                                & vessel segmentation in retina                                \\
			Structured Analysis of the Retina\cite{hoover2000locating}                                    & STARE           & 20                                                                                & vessel segmentation in retina                                \\
			Child Heart and Health Study in England Database\cite{owen2009measuring}                      & CHDB            & 14                                                                                & blood vessel segmentation in retina                           \\
			High Resolution Fundus\cite{odstrcilik2013retinal}                                            & HRF             & 45                                                                                & vessel segmentation in retina                                \\
			Kaggle Retinopathy Detection Challenge 2015\cite{graham2015kaggle}                            & KR15            & ---\footref{publicdatabaselabel}                                                  & diabetic retinopathy classification                          \\
			TeleOptha\cite{decenciere2013teleophta}                                                       & e-optha         & 381                                                                               & MA and hemorrhage detection                                  \\
			Messidor\cite{decenciere2014feedback}                                                         & Messidor        & 1200                                                                              & diabetic retinopathy diagnosis                               \\
			Messidor2\cite{decenciere2014feedback}                                                        & Messidor2       & 874                                                                               & diabetic retinopathy diagnosis                               \\
			Diaretdb1\cite{kauppi2013constructing}                                                        & DIA             & 89                                                                                & MA and hemorrhage detection                                  \\
			Retinopathy Online Challenge\cite{niemeijer2010retinopathy}                                   & ROC             & 100                                                                               & MA detection                                                 \\
			IOSTAR\cite{zhang2016robust}                                                                  & IOSTAR          & 30                                                                                & vessel segmentation in retina using SLO                      \\
			RC-SLO\cite{zhang2016robust}                                                                  & RC-SLO          & 40                                                                                & vessel segmentation in retina using SLO                      \\
			\midrule
			\multicolumn{4}{l}{\thead{Other Imaging Databases}}                                                                                                                                                                                                                \\
			\midrule
			MICCAI 2011 Lumen+External Elastic Laminae\cite{balocco2014standardized}                      & IV11            & 32                                                                                & lumen and external contour segmentation in IVUS              \\
			UK Biobank\cite{sudlow2015uk}                                                                 & UKBDB           & ---\footref{publicdatabaselabel}                                                  & multiple imaging databases                                   \\
			Coronary Artery Stenoses Detection and Quantification\cite{kiricsli2013standardized}          & CASDQ           & 48                                                                                & cardiac CT angiography for coronary artery stenoses          \\
			\midrule
			\multicolumn{4}{l}{\thead{Multimodal Databases}}                                                                                                                                                                                                                   \\
			\midrule
			VORTAL\cite{charlton2016assessment}                                                           & VORTAL          & 45                                                                                & respiratory rate estimation with ECG and PCG                 \\
			Left Atrium Segmentation Challenge STACOM 2013\cite{tobon2015benchmark}                       & STA13           & 30                                                                                & left atrium segmentation with MRI, CT                        \\
			MICCAI MMWHS 2017\cite{zhuang2016multi}                                                       & MM17            & 60                                                                                & 120 images for whole heart segmentation with MRI, CT         \\
			\bottomrule
		\end{tabularx}
	\end{minipage}
\end{table*}

\section{Deep learning using structured data}
\label{sec:structured}
Structured data mainly include EHRs and exist in an organized form based on data fields typically held in relational databases.
A summary of deep learning applications using structured data is shown in Table~\ref{table:structured}.

RNNs have been used for cardiovascular disease diagnosis using structured data.
In\cite{gopalswamy2017deep} the authors predict the BP during surgery and length of stay after the surgery using LSTM\@.
They performed experiments on a dataset of 12036 surgeries that contain information on intraoperative signals (body temperature, respiratory rate, heart rate, DBP, SBP, fraction of inspired $O_2$ and end-tidal $CO_2$), achieving better results than KNN and SVM baselines.
Choi et al.\cite{choi2016using} trained a GRU with longitudinal EHR data, detecting relations among time-stamped events (disease diagnosis, medication orders, etc.) using an observation window.
They diagnose Heart Failure (HF) achieving AUC 0.777 for a 12 month window and 0.883 for 18 month window, higher than the MLP, SVM and KNN baselines.
Purushotham et al.\cite{purushotham2018benchmarking} compared the super learner (ensemble of shallow machine learning algorithms)\cite{polley2010super} with FNN, RNN and a multimodal deep learning model proposed by the authors on the MIMIC database.
The proposed framework uses FNN and GRU for handling non-temporal and temporal features respectively, thus learning their shared latent representations for prediction.
The results show that deep learning methods consistently outperform the super learner in the majority of the prediction tasks of the MIMIC (predictions of in-hospital mortality AUC 0.873, short-term mortality AUC 0.871, long-term mortality AUC 0.87 and ICD-9 code AUC 0.777).
Kim et al.\cite{kim2017highrisk} created two medical history prediction models using attention networks and evaluated them on 50000 hypertension patients.
They showed that using a bi-directional GRU-based model provides better discriminative capability than the convolutional-based model which has shorter training time with competitive accuracy.

AEs have been used for cardiovascular disease diagnosis using structured data.
Hsiao et al.\cite{hsiao2016deep} trained an AE and a softmax layer for risk analysis of four categories of cardiovascular diseases.
The input included demographics, ICD-9 codes from outpatient records and concentration pollutants, meteorological parameters from environmental records.
Huang et al.\cite{huang2018regularized} trained a SDAE using an EHR dataset of 3464 patients to predict ACS\@.
The SDAE has two regularization constraints that makes reconstructed feature representations contain more risk information thus capturing characteristics of patients at similar risk levels, and preserving the discriminating information across different risk levels.
Then, they append a softmax layer, which is tailored to the clinical risk prediction problem.

DBNs have also been used in combination with structured data besides RNNs and AEs.
In\cite{kim2017statistics} the authors first performed a statistical analysis of a dataset with 4244 records to find variables related to cardiovascular disease from demographics and lifestyle data (age, gender, cholesterol, high-density lipoprotein, SBP, DBP, smoking, diabetes).
Then, they developed a DBN model for predicting cardiovascular diseases (hypertension, hyperlipidemia, Myocardical Infarction (MI), angina pectoris).
They compared their model with Naive Bayes, Logistic Regression, SVM, RF and a baseline DBN achieving better results.

According to the literature, RNNs are widely used in cardiology structured data because they are capable in finding optimal temporal features better than other deep/machine learning methods.
On the other hand, applications in this area are relatively few and this is mainly because there is a small number of public databases in this area which prevents further evaluation and comparison of different architectures on these datasets.
Additionally, structured databases by their design contain less information for an individual patient and focus more on groups of patients, making them more suitable for epidemiologic studies rather than cardiology.

\begin{table*}[!t]
	\caption{Deep learning applications using structured data}
	\label{table:structured}
	\begin{minipage}{\textwidth}
		\centering
		\begin{tabularx}{\textwidth}{l c l l}
			\toprule
			\thead{Reference}                                  & \thead{Method} & \thead{Application/Notes\footnote{In parenthesis the databases used.}}                        & \thead{Result\footnote{\label{structuredlabel}There is a wide variability in results reporting. All results are accuracies besides~\cite{choi2016using} which report AUC and~\cite{hsiao2016deep} which is a statistical study.}} \\
			\midrule
			Gopalswamy 2017\cite{gopalswamy2017deep}           & LSTM           & predict BP and length of stay using multiple vital signals (private)                          & 73.1\%                                                                                                                                                                                                                                                                                                                                                                                   \\
			Choi 2016\cite{choi2016using}                      & GRU            & predict initial diagnosis of HF using a GRU with an observation window (private)              & 0.883\footref{structuredlabel}                                                                                                                                                                                                                                                                                                                                                          \\
			Purushotham 2018\cite{purushotham2018benchmarking} & FNN, GRU       & prediction tasks of MIMIC using a FNN and a GRU-based network (MIMIC)                           & \textit{multiple}                                                                                                                                                                                                                                                                                                                                                          \\
			Kim 2017\cite{kim2017highrisk}                     & GRU, CNN       & predict onset of high risky vascular disease using a bidirectional GRU and a 1D CNN (private) & \textit{multiple}                                                                                                                                                                                                                                                                                                                                                          \\
			Hsiao 2016\cite{hsiao2016deep}                     & AE             & analyze CVD risk using an AE and a softmax on outpatient and meteorological dataset (private) & ---\footref{structuredlabel}                                                                                                                                                                                                                                                                                                                                                             \\
			Kim 2017\cite{kim2017statistics}                   & DBN            & predict cardiovascular risk using a DBN (KNH)                                                 & 83.9\%                                                                                                                                                                                                                                                                                                                                                                                   \\
			Huang 2018\cite{huang2018regularized}              & SDAE           & predict ACS risk using a SDAE with two reguralization constrains and a softmax (private)      & 73.0\%                                                                                                                                                                                                                                                                                                                                                                                   \\
			\bottomrule
		\end{tabularx}
	\end{minipage}
\end{table*}

\section{Deep learning using signals}
\label{sec:signals}
Signal modalities include time-series such as Electrocardiograms (ECGs), Phonocardiograms (PCGs), oscillometric and wearable data.
One reason that traditional machine learning have worked sufficiently well in this area in previous years is because of the use of handcrafted and carefully designed features by experts such as statistical measures from the ECG beats and the RR interval\cite{faziludeen2013ecg}.
Deep learning can improve results when the annotations are noisy or when it is difficult to manually create a model.
A summary of deep learning applications using signals is shown in Tables~\ref{table:signals1} and~\ref{table:signals2}.

\begin{table*}[!t]
	\caption{Deep learning applications using ECG}
	\label{table:signals1}
	\begin{minipage}{\textwidth}
		\centering
		\begin{tabularx}{\textwidth}{l c l l}
			\toprule
			\thead{Reference}                              & \thead{Method} & \thead{Application/Notes\footnote{In parenthesis the databases used by paper or by papers in subsection.}} & \thead{Accuracy\footnote{\label{signals1label}There is a wide variability in results reporting. The results of~\cite{kiranyaz2016real} is for ventricular/supraventricular ectopic beats,~\cite{isin2017cardiac} is for three types of arrhythmias,~\cite{wu2016novel} is for five types of arrhythmias,~\cite{rajpurkar2017cardiologist} report precision,~\cite{xiong2015denoising} report SNR and multiple results depending on added noise, the result of~\cite{taji2017false} is without noise in a noise resilience study,~\cite{xiao2018monitoring} report AUC,~\cite{al2016deep} report multiple accuracies for supraventricular/ventricular ectopic beats,~\cite{wu2016myocardial} report sensitivity and specificity,~\cite{hwang2018deep} report results for two cases.}} \\
			\midrule
			\multicolumn{4}{l}{\thead{arrhythmia detection (MITDB)}}                                                                                                                                                                                                                                                                                                                                                                                                                                                                                                                                                                                                                                                                                                                                                                                                                                                                                                                                                                                                                                \\
			\midrule
			Zubair 2016\cite{zubair2016automated}          & CNN            & non-linear transform for R-peak detection and a 1D CNN with a variable learning rate                       & 92.7\%                                                                                                                                                                                                                                                                                                                                                                                                                                                                                                                                                                                                                                                                                                                                                                                                                                                                                                   \\
			Li 2017\cite{li2017classification}             & CNN            & WT for denoising and R-peak detection and a two \textit{layer} 1D CNN                                      & 97.5\%                                                                                                                                                                                                                                                                                                                                                                                                                                                                                                                                                                                                                                                                                                                                                                                                                                                                                                   \\
			Kiranyaz 2016\cite{kiranyaz2016real}           & CNN            & patient-specific CNN using adaptive 1D convolutional layers                                                & 99\%,97.6\%\footref{signals1label}                                                                                                                                                                                                                                                                                                                                                                                                                                                                                                                                                                                                                                                                                                                                                                                                                                                                       \\
			Isin 2017\cite{isin2017cardiac}                & CNN            & denoising filters, Pan-Tomkins, AlexNet for feature extraction and PCA for classification                  & 92.0\%\footref{signals1label}                                                                                                                                                                                                                                                                                                                                                                                                                                                                                                                                                                                                                                                                                                                                                                                                                                                                            \\
			Luo 2017\cite{luo2017patient}                  & SDAE           & denoising filters, derivative-based R-peak detection, WT, SDAE and softmax                                 & 97.5\%                                                                                                                                                                                                                                                                                                                                                                                                                                                                                                                                                                                                                                                                                                                                                                                                                                                                                                   \\
			Jiang 2017\cite{jiang2017heartbeat}            & SDAE           & denoising filters, Pan-Tomkins, SDAE and FNN                                                               & 97.99\%                                                                                                                                                                                                                                                                                                                                                                                                                                                                                                                                                                                                                                                                                                                                                                                                                                                                                                  \\
			Yang 2017\cite{yang2017novel}                  & SSAE           & normalize ECG, fine-tuned SSAE                                                                             & 99.45\%                                                                                                                                                                                                                                                                                                                                                                                                                                                                                                                                                                                                                                                                                                                                                                                                                                                                                                  \\
			Wu 2016\cite{wu2016novel}                      & DBN            & denoising filters, ecgpuwave, two types of RBMs                                                            & 99.5\%\footref{signals1label}                                                                                                                                                                                                                                                                                                                                                                                                                                                                                                                                                                                                                                                                                                                                                                                                                                                                            \\
			\midrule
			\multicolumn{4}{l}{\thead{arrhythmia detection}}                                                                                                                                                                                                                                                                                                                                                                                                                                                                                                                                                                                                                                                                                                                                                                                                                                                                                                                                                                                                                                        \\
			\midrule
			Wu 2018\cite{wu2018personalizing}              & CNN            & active learning and a two \textit{layer} CNN fed with ECG and RR interval (MITDB, DeepQ)                   & \textit{multiple}                                                                                                                                                                                                                                                                                                                                                                                                                                                                                                                                                                                                                                                                                                                                                                                                                                                                         \\
			Rajpurkar 2017\cite{rajpurkar2017cardiologist} & CNN            & 34-layer CNN (private wearable dataset)                                                                    & 80\%\footref{signals1label}                                                                                                                                                                                                                                                                                                                                                                                                                                                                                                                                                                                                                                                                                                                                                                                                                                                                              \\
			Acharya 2017\cite{acharya2017automateda}       & CNN            & four \textit{layer} CNN (AFDB, MITDB, Creighton)                                                           & 92.5\%                                                                                                                                                                                                                                                                                                                                                                                                                                                                                                                                                                                                                                                                                                                                                                                                                                                                                                   \\
			Schwab 2017\cite{schwab2017beat}               & RNN            & ensemble of RNNs with an attention mechanism (PHY17)                                                       & 79\%                                                                                                                                                                                                                                                                                                                                                                                                                                                                                                                                                                                                                                                                                                                                                                                                                                                                                                     \\
			\midrule
			\multicolumn{4}{l}{\thead{AF detection}}                                                                                                                                                                                                                                                                                                                                                                                                                                                                                                                                                                                                                                                                                                                                                                                                                                                                                                                                                                                                                                                \\
			\midrule
			Yao 2017\cite{yao2017atrial}                   & CNN            & multiscale CNN (AFDB, LTAFDB, private)                                                                     & 98.18\%                                                                                                                                                                                                                                                                                                                                                                                                                                                                                                                                                                                                                                                                                                                                                                                                                                                                                                  \\
			Xia 2018\cite{xia2018detecting}                & CNN            & CNN with spectrograms from short time fourier transform or stationary WT (AFDB)                            & 98.29\%                                                                                                                                                                                                                                                                                                                                                                                                                                                                                                                                                                                                                                                                                                                                                                                                                                                                                                  \\
			Andersen 2018\cite{andersen2018deep}           & CNN, LSTM      & RR intervals with a CNN-LSTM network (MITDB, AFDB, NSRDB)                                                  & 87.40\%                                                                                                                                                                                                                                                                                                                                                                                                                                                                                                                                                                                                                                                                                                                                                                                                                                                                                                  \\
			Xiong 2015\cite{xiong2015denoising}            & AE             & scale-adaptive thresholding WT and a denoising AE (MITDB, NSTDB)                                           & $\sim$18.7\footref{signals1label}                                                                                                                                                                                                                                                                                                                                                                                                                                                                                                                                                                                                                                                                                                                                                                                                                                                                        \\
			Taji 2017\cite{taji2017false}                  & DBN            & false alarm reduction during AF detection in noisy ECG signals (AFDB, NSTDB)                               & 87\%\footref{signals1label}                                                                                                                                                                                                                                                                                                                                                                                                                                                                                                                                                                                                                                                                                                                                                                                                                                                                              \\
			\midrule
			\multicolumn{4}{l}{\thead{Other tasks}}                                                                                                                                                                                                                                                                                                                                                                                                                                                                                                                                                                                                                                                                                                                                                                                                                                                                                                                                                                                                                                                 \\
			\midrule
			Xiao 2018\cite{xiao2018monitoring}             & CNN            & classify ST events from ECG using transfer learning on Inception v3 (LTSTDB)                               & 0.867\footref{signals1label}                                                                                                                                                                                                                                                                                                                                                                                                                                                                                                                                                                                                                                                                                                                                                                                                                                                                           \\
			Rahhal 2016\cite{al2016deep}                   & SDAE           & SDAE with sparsity constrain and softmax (MITDB, INDB, SVDB)                                               & \textgreater{99\%}\footref{signals1label}                                                                                                                                                                                                                                                                                                                                                                                                                                                                                                                                                                                                                                                                                                                                                                                                                                                                \\
			Abrishami 2018\cite{abrishami2018p}            & Multiple       & compared a FNN, a CNN and a CNN with dropout for ECG wave localization (QTDB)                              & 96.2\%                                                                                                                                                                                                                                                                                                                                                                                                                                                                                                                                                                                                                                                                                                                                                                                                                                                                                                   \\
			Wu 2016\cite{wu2016myocardial}                 & SAE            & detect and classify MI using a SAE and multi-scale discrete WT (PTBDB)                                     & $\sim$99\%\footref{signals1label}                                                                                                                                                                                                                                                                                                                                                                                                                                                                                                                                                                                                                                                                                                                                                                                                                                                                        \\
			Reasat 2017\cite{reasat2017detection}          & Inception      & detect MI using Inception block for each ECG lead (PTBDB)                                                  & 84.54\%                                                                                                                                                                                                                                                                                                                                                                                                                                                                                                                                                                                                                                                                                                                                                                                                                                                                                                  \\
			Zhong 2018\cite{zhong2018deep}                 & CNN            & three \textit{layer} CNN for classifying fetal ECG segments (PHY13)                                        & 77.85\%                                                                                                                                                                                                                                                                                                                                                                                                                                                                                                                                                                                                                                                                                                                                                                                                                                                                                                  \\
			\midrule
			\multicolumn{4}{l}{\thead{Other tasks (private databases)}}                                                                                                                                                                                                                                                                                                                                                                                                                                                                                                                                                                                                                                                                                                                                                                                                                                                                                                                                                                                                                             \\
			\midrule
			Ripoll 2016\cite{ripoll2016ecg}                & RBM            & identify abnormal ECG using pretrained RBMs                                                                & 85.52\%                                                                                                                                                                                                                                                                                                                                                                                                                                                                                                                                                                                                                                                                                                                                                                                                                                                                                                  \\
			Jin 2017\cite{jin2017classification}           & CNN            & identify abnormal ECG using lead-CNN and rule inference                                                    & 86.22\%                                                                                                                                                                                                                                                                                                                                                                                                                                                                                                                                                                                                                                                                                                                                                                                                                                                                                                  \\
			Liu 2018\cite{liu2018detecting}                & Multiple       & compared Inception and a 1D CNN for premature ventricular contraction in ECG                               & 88.5\%                                                                                                                                                                                                                                                                                                                                                                                                                                                                                                                                                                                                                                                                                                                                                                                                                                                                                                   \\
			Hwang 2018\cite{hwang2018deep}                 & CNN, RNN       & detect stress with an one convolutional layer with dropout and two RNNs on                                 & 87.39\%\footref{signals1label}                                                                                                                                                                                                                                                                                                                                                                                                                                                                                                                                                                                                                                                                                                                                                                                                                                                                           \\
			\bottomrule
		\end{tabularx}
	\end{minipage}
\end{table*}

\subsection{Electrocardiogram}
ECG is the method of measuring the electrical potentials of the heart to diagnose heart related problems\cite{badnjevic2017inspection}.
It is non-invasive, easy to acquire and provides a useful proxy for disease diagnosis.
It has mainly been used for arrhythmia detection utilizing the large number of publicly available ECG databases as shown in Table~\ref{table:cardiologypublicdatabases}.

\subsubsection{Arrhythmia detection with MITDB}
CNNs have been used for arrhythmia detection with MITDB\@.
Zubair et al.\cite{zubair2016automated} detected the R-peak using a non-linear transformation and formed a beat segment around it.
Then, they used the segments to train a three \textit{layer} 1D CNN with variable learning rate depending on the mean square error and achieved better results than previous state-of-the-art.
Li et al.\cite{li2017classification} used WT to remove high frequency noise and baseline drift and biorthogonal spline wavelet for detecting the R-peak.
Then, they created and resampled segments around the R-peak before feeding them to a two \textit{layer} 1D CNN\@.
In their article Kiranyaz et al.\cite{kiranyaz2016real} trained patient-specific CNNs that can be used to classify long ECG data stream or for real-time ECG monitoring and early alert system on a wearable device.
The CNN consisted of three \textit{layers} of an adaptive implementation of 1D convolution layers.
They achieved 99\% and 97.6\% in classifying ventricular and supraventricular ectopic beats respectively.
In\cite{isin2017cardiac} the authors used mean removal for dc removal, moving average filter for high frequency removal, derivative-based filter for baseline wander removal and a comb filter for power line noise removal.
They detected QRS with Pan-Tompkins algorithm\cite{pan1985real}, extracted segments using samples after the R-peak and converted them to $256\times 256\times 3$ binary images.
The images were then fed to an AlexNet feature extractor trained on ImageNet and then to Principal Component Analysis (PCA).
They achieved high accuracy classifying three types of arrhythmias of MITDB\@.

AEs have also been used for arrhythmia detection with MITDB\@.
In their article Luo et al.\cite{luo2017patient} utilized quality assessment to remove low quality heartbeats, two median filters for removing power line noise, high-frequency noise and baseline drift.
Then, they used a derivative-based algorithm to detect R-peaks and time windows to segment each heartbeat.
Modified frequency slice WT was used to calculate the spectrogram of each heartbeat and a SDAE for extracting features from the spectrogram.
Then, they created a classifier for four arrhythmias from the encoder of the SDAE and a softmax, achieving an overall accuracy of 97.5\%.
In\cite{jiang2017heartbeat} the authors denoised the signals with a low-pass, a bandstop and a median filter.
They detected R-peaks using Pan-Tomkins algorithm and segmented/resampled the heartbeats.
Features were extracted from the heartbeat signal using a SDAE, and a FNN was used to classify the heartbeats in 16 types of arrhythmia.
Comparable performance with previous methods based on feature engineering was achieved.
Yang et al.\cite{yang2017novel} normalized the ECG and then fed it to a Stacked Sparse AE (SSAE) which they fine-tuned.
They classify on six types of arrhythmia achieving accuracy of 99.5\% while also demonstrating the noise resilience of their method with artificially added noise.

DBNs have also been used for this task besides CNNs and AEs.
Wu et al.\cite{wu2016novel} used median filters to remove baseline wander, a low-pass filter to remove power-line and high frequency noise.
They detected R-peaks using ecgpuwave software from Physionet and segmented and resampled ECG beats.
Two types of RBMs, were trained for feature extraction from ECG for arrhythmia detection.
They achieved 99.5\% accuracy on five classes of MITDB\@.

\subsubsection{Arrhythmia detection with other databases}
CNNs have been used for arrhythmia detection using other databases besides solely on MITDB\@.
In\cite{wu2018personalizing} the authors created a two \textit{layer} CNN using the DeepQ\cite{wu2017deepq} and MITDB to classify four arrhythmias types.
The signals are heavily preprocessed with denoising filters (median, high-pass, low-pass, outlier removal) and they are segmented to 0.6 seconds around the R-peak.
Then, they are fed to the CNN along with the RR interval for training.
The authors also employ an active learning method to achieve personalized results and improved precision, achieving high sensitivity and positive predictivity in both datasets.
Rajpurkar et al.\cite{rajpurkar2017cardiologist} created an ECG wearable dataset that contains the largest number of unique patients (30000) than previous datasets and used it to train a 34-layer residual-based CNN\@.
Their model detects a wide range of arrhythmias a total of 14 output classes, outperforming the average cardiologist.
In their article Acharya et al.\cite{acharya2017automateda} trained a four \textit{layer} CNN on AFDB, MITDB and CREI, to classify between normal, AF, atrial flutter and ventricular fibrillation.
Without detecting the QRS they achieved comparable performance with previous state-of-the-art methods that were based on R-peak detection and feature engineering.
The same authors have also trained the previous CNN architecture for identifying shockable and non-shockable ventricular arrhythmias\cite{acharya2018automated}, identify CAD patients with FAN and INDB\cite{acharya2017automatedb}, classify CHF with CHFDB, NSTDB, FAN\cite{acharya2018deep} and also tested its noise resistance with WT denoising\cite{acharya2017application}.

An application of RNNs in this area is from Schwab et al.\cite{schwab2017beat} that built an ensemble of RNNs that distinguishes between normal sinus rhythms, AF, other types of arrhythmia and noisy signals.
They introduced a task formulation that segments ECG into heartbeats to reduce the number of time steps per sequence.
They also extended the RNNs with an attention mechanism that enables them to reason which heartbeats the RNNs focus on to make their decisions and achieved comparable to state-of-the-art performance using fewer parameters than previous methods.

\subsubsection{AF detection}
CNNs have been used for AF detection.
Yao et al.\cite{yao2017atrial} extracted instant heart rate sequence, which is fed to an end-to-end multi-scale CNN that outputs the AF detection result, achieving better results than previous methods in terms of accuracy.
Xia et al.\cite{xia2018detecting} compared two CNNs, with three and two \textit{layers}, that were fed with spectrograms of signals from AFDB using Short-Term Fourier Transform and stationary WT respectively.
Their experiments concluded that the use of stationary WT achieves a slightly better accuracy for this task.

Besides CNNs other architectures have been used for AF detection.
Andersen et al.\cite{andersen2018deep} converted ECG signals from AFDB, into RR intervals to classify them for AF detection.
Then, they segmented the RR intervals to 30 samples each and fed them to a network with two \textit{layers} followed by a pooling layer and a LSTM layer with 100 units.
The method was validated on MITDB and NSRDB achieving an accuracy that indicates its generalizability.
In\cite{xiong2015denoising} the authors added noise signals from the NSTDB to the MITDB and then used scale-adaptive thresholding WT to remove most of the noise and a denoised AE to remove the residual noise.
Their experiments indicated that increasing the number of training data to 1000 the signal-to-noise ratio increases dramatically after denoising.
Taji et al.\cite{taji2017false} trained a DBN to classify acceptable from unacceptable ECG segments to reduce the false alarm rate caused by poor quality ECG during AF detection.
Eight different levels of ECG quality are provided by contaminating ECG with motion artifact from the NSTDB for validation.
With an SNR of $-20dB$ in the ECG signal their method achieved an increase of 22\% in accuracy compared to the baseline model.

\subsubsection{Other tasks with public databases}
ECG beat classification was also performed by a number of studies using public databases.
In\cite{xiao2018monitoring} the authors finetuned a Inception v3 trained on ImageNet, using signals from LTSTDB for classifying ST events.
The training samples were over 500000 segments of ST and non-ST ECG signals with ten second duration that were converted to images.
They achieve comparable performance with previous complex rule-defined methods.
Rahhal et al.\cite{al2016deep} trained a SDAEs with sparsity constraint and a softmax for ECG beat classification.
At each iteration the expert annotates the most uncertain ECG beats in the test set, which are then used for training, while the output of the network assigns the confidence measures to each test beat.
Experiments performed on MITDB, INDB, SVDB indicate the robustness and computational efficiency of the method.
In\cite{abrishami2018p} the authors trained three separate architectures to identify the P-QRS-T waves in ECG with QTDB\@.
They compared a two layer FNN, a two \textit{layer} CNN and a two \textit{layer} CNN with dropout with the second one achieving the best results.

ECG has also been used for MI detection and classification.
In their article Wu et al.\cite{wu2016myocardial} detected and classified MI with PTBDB\@.
They used multi-scale discrete WT to facilitate the extraction of MI features at specific frequency resolutions and softmax regression to build a multi-class classifier based on the learned features.
Their validation experiments show that their method performed better than previous methods in terms of sensitivity and specificity.
PTBDB was also used by Reasat et al.\cite{reasat2017detection} to train an inception-based CNN\@.
Each ECG lead is fed to an inception block, followed by concatenation, global average pooling and a softmax.
The authors compared their method with a previous state-of-the-art method that uses SWT, demonstrating better results.

Fetal QRS complexes were identified by a three \textit{layer} CNN with dropout by Zhong et al.\cite{zhong2018deep} with PHY13.
First, the bad quality signals are discarded using sample entropy and then normalized segments with duration of 100ms are fed to the CNN for training.
The authors compared their method with KNN, Naive Bayes and SVM achieving significantly better results.

\subsubsection{Other tasks with private databases}
Abnormal ECG detection was studied by a number of papers.
Ripoll et al.\cite{ripoll2016ecg} used RBM-based pretrained models with ECGs from 1390 patients to assess whether a patient from ambulatory care or emergency should be referred to a cardiology service.
They compared their model with KNN, SVM, extreme learning machines and an expert system achieving better results in accuracy and specificity.
In\cite{jin2017classification} the authors train a model that classifies normal and abnormal subjects with 193690 ECG records of 10 to 20 seconds.
Their model consisted of two parallel parts; the statistical learning and a rule inference.
In statistical learning the ECGs are preprocessed using bandpass and lowpass filters, then fed to two parallel lead-CNNs and finally Bayesian fusion is employed to combine the probability outputs.
In rule inference, the R-peak positions in the ECG record are detected and four disease rules are used for analysis.
Finally, they utilize bias-average to determine the result.

Other tasks include premature ventricular contraction classification and stress detection.
Liu et al.\cite{liu2018detecting} used a single lead balanced dataset of 2400 normal and premature ventricular contraction ECGs from Children's Hospital of Shanghai for training.
Two separate models were trained using the waveform images.
The first one was a two \textit{layer} CNN with dropout and the second an Inception v3 trained on Imagenet.
Another three models were trained using the signals as 1D.
The first model was a FNN with dropout, the second a three \textit{layer} 1D CNN and the third a 2D CNN same as the first but trained with a stacked version of the signal (also trained with data augmentation).
Experiments by the authors showed that the three \textit{layer} 1D CNN created better and more stable results.
In\cite{hwang2018deep} the authors trained a network with an one convolutional layer with dropout followed by two RNNs to identify stress using short-term ECG data.
They showed that their network achieved the best results compared with traditional machine learning methods and baseline DNNs.

\subsubsection{Overall view on deep learning using ECG}
Many deep learning methods have used ECG to train deep learning models utilizing the large number of databases that exist for that modality.
It is evident from the literature that most deep learning methods (mostly CNNs and SDAEs) in this area consist of three parts: filtering for denoising, R-peak detection for beat segmentation and the neural network for feature extraction.
Another popular set of methods is the conversion of ECGs to images, to utilize the wide range of architectures and pretrained models that have already been built for imaging modalities.
This was done using spectrogram techniques\cite{luo2017patient, xia2018detecting} and conversion to binary image\cite{xiao2018monitoring, liu2018detecting, isin2017cardiac}.

\subsection{Phonocardiogram with Physionet 2016 Challenge}
Physionet/Computing in Cardiology (Cinc) Challenge 2016 (PHY16) was a competition for classification of normal/abnormal heart sound recordings.
The training set consists of five databases (A through E) that contain 3126 Phonocardiograms (PCGs), lasting from 5 seconds to 120 seconds.

\begin{table*}[!t]
	\caption{Deep learning applications using PCG and other signals}
	\label{table:signals2}
	\begin{minipage}{\textwidth}
		\centering
		\begin{tabularx}{\textwidth}{l c l l}
			\toprule
			\thead{Reference}                           & \thead{Method} & \thead{Application/Notes\footnote{In parenthesis the databases used by paper or by papers in subsection. In the `PCG/Physionet 2016 Challenge' subtable all papers use PHY besides\cite{chen2017s1}, and in the `Other signals' subtable all papers use private databases besides~\cite{poh2018diagnostic}.}} & \thead{Accuracy\footnote{\label{signals2label}There is a wide variability in results reporting.~\cite{kucharski2017deep} report specificity,~\cite{pan2017variation} report results for SBP and DBP,~\cite{gotlibovych2018end} report sensitivity, specificity,~\cite{poh2018diagnostic} report positive predictive value,~\cite{ballinger2018deepheart} report AUC for diabetest, results are also reported for high cholesterol sleep apnea and high BP.}} \\
			\midrule
			\multicolumn{4}{l}{\thead{PCG/Physionet 2016 Challenge}}                                                                                                                                                                                                                                                                                                                                                                                                                                                                       \\
			\midrule
			Rubin 2017\cite{rubin2017recognizing}       & CNN            & logistic regression hidden-semi markov, MFCCs and a two \textit{layer} CNN                                 & 83.99\%                                                                                                                                                                                                                                                                                                                                                                                                  \\
			Kucharski 2017\cite{kucharski2017deep}      & CNN            & spectrogram and five \textit{layer} CNN with dropout                                                       & 91.6\%\footref{signals1label}                                                                                                                                                                                                                                                                                                                                                                            \\
			Dominguez 2018\cite{dominguez2018deep}      & CNN            & spectrogram and modified AlexNet                                                                           & 94.16\%                                                                                                                                                                                                                                                                                                                                                                                                  \\
			Potes 2016\cite{potes2016ensemble}          & CNN            & ensemble of Adaboost and CNN and outputs combined with decision rule                                       & 86.02\%                                                                                                                                                                                                                                                                                                                                                                                                  \\
			Ryu 2016\cite{ryu2016classification}        & CNN            & denoising filters and four \textit{layer} CNN                                                              & 79.5\%                                                                                                                                                                                                                                                                                                                                                                                                   \\
			Chen 2017\cite{chen2017s1}                  & DBN            & recognize S1 and S2 heart sounds using MFCCs, K-means and DBN (private)                                    & 91\%                                                                                                                                                                                                                                                                                                                                                                                                     \\
			\midrule
			\multicolumn{4}{l}{\thead{Other signals}}                                                                                                                                                                                                                                                                                                                                                                                                                                                                                                                                            \\
			\midrule
			Lee 2017\cite{lee2017deepa}                 & DBN            & estimate BP using bootstrap-aggregation, Monte-Carlo and DBN with oscillometry data                        & \textit{multiple}                                                                                                                                                                                                                                                                                                                                                                         \\
			Pan 2017\cite{pan2017variation}             & CNN            & assess Korotkoff sounds using a three \textit{layer} CNN with oscillometry data                            & \textit{multiple}\footref{signals1label}                                                                                                                                                                                                                                                                                                                                                                        \\
			Shashikumar 2017\cite{shashikumar2017deep}  & CNN            & detect AF using ECG, photoplethysmography, accelerometry with WT and a CNN                                 & 91.8\%                                                                                                                                                                                                                                                                                                                                                                                                   \\
			Gotlibovych 2018\cite{gotlibovych2018end}   & CNN, LSTM      & detect AF using PPG from wearable and a LSTM-based CNN                                                     & \textgreater{99\%}\footref{signals1label}                                                                                                                                                                                                                                                                                                                                                                \\
			Poh 2018\cite{poh2018diagnostic}            & CNN            & detect four rhythms on PPG using a densely CNN (MIMIC, VORTAL, PPGDB)                                      & 87.5\%\footref{signals1label}                                                                                                                                                                                                                                                                                                                                                                            \\
			Ballinger 2018\cite{ballinger2018deepheart} & LSTM           & predict diabetes, high cholesterol, high BP, and sleep apnoea using sensor data and LSTM                   & 0.845\footref{signals1label}                                                                                                                                                                                                                                                                                                                                                                           \\
			\bottomrule
		\end{tabularx}
	\end{minipage}
\end{table*}

Most of the methods convert PCGs to images using spectrogram techniques.
Rubin et al.\cite{rubin2017recognizing} used a logistic regression hidden semi-markov model for segmenting the start of each heartbeat which then were transformed into spectrograms using Mel-Frequency Cepstral Coefficients (MFCCs).
Each spectrogram was classified into normal or abnormal using a two \textit{layer} CNN which had a modified loss function that maximizes sensitivity and specificity, along with a regularization parameter.
The final classification of the signal was the average probability of all segment probabilities.
They achieved an overall score of 83.99\% placing eighth at PHY16 challenge.
Kucharski et al.\cite{kucharski2017deep} used an eight second spectrogram on the segments, before feeding them to a five \textit{layer} CNN with dropout.
Their method achieved 99.1\% sensitivity and 91.6\% specificity which are comparable to state-of-the-art methods on the task.
Dominguez et al.\cite{dominguez2018deep} segmented the signals and preprocessed them using the neuromorphic auditory sensor\cite{jimenez2017binaural} to decompose the audio information into frequency bands.
Then, they calculated the spectrograms which were fed to a modified version of AlexNet.
Their model achieved accuracy of 94.16\% a significant improvement compared with the winning model of PHY16.
In\cite{potes2016ensemble} the authors used Adaboost which was fed with spectrogram features from PCG and a CNN which was trained using cardiac cycles decomposed into four frequency bands.
Finally, the outputs of the Adaboost and the CNN were combined to produce the final classification result using a simple decision rule.
The overall accuracy was 89\%, placing this method first in the official face of PHY16.

Models that did not convert the PCGs to spectrograms seemed to have lesser performance.
Ryu et al.\cite{ryu2016classification} applied Window-sinc Hamming filter for denoising, scaled the signal and used a constant window for segmentation.
They trained a four \textit{layer} 1D CNN using the segments and the final classification was the average of all segment probabilities.
An overall accuracy of 79.5\% was achieved in the official phase of PHY16.

Phonocardiograms have also been used for tasks such as S1 and S2 heart sound recognition by Chen et al.\cite{chen2017s1}.
They converted heart sound signals into a sequence of MFCCs and then applied K-means to cluster the MFCC features into two groups to refine their representation and discriminative capability.
The features are then fed to a DBN to perform S1 and S2 classification.
The authors compared their method with KNN, Gaussian mixture models, logistic regression and SVM performing the best results.

According to the literature, CNNs are the majority of neural network architectures used for solving tasks with PCG\@.
Moreover, just like in ECG, many deep learning methods converted the signals to images using spectrogram techniques\cite{potes2016ensemble, rubin2017recognizing, kucharski2017deep, dominguez2018deep, pan2017variation, shashikumar2017deep}.

\subsection{Other signals}
\subsubsection{Oscillometric data}
Oscillometric data are used for estimating SBP and DBP which are the haemodynamic pressures exerted within the arterial system during systole and diastole respectively\cite{everly2012clinical}.

DBNs have been used for SBP and DBP estimation.
In their article Lee et al.\cite{lee2017deepa} used bootstrap-aggregation to create ensemble parameters and then employed Adaboost to estimate SBP and DBP\@.
Then, they used bootstrap and Monte-Carlo in order to determine the confidence intervals based on the target BP, which was estimated using the DBN ensemble regression estimator.
This modification greatly improved the BP estimation over the baseline DBN model.
Similar work has been done on this task by the same authors in\cite{lee2017oscillometric, lee2017deepc, lee2017deepd}.

Oscillometric data have also been used by Pan et al.\cite{pan2017variation} for assessing the variation of Korotkoff sounds.
The beats were used to create windows centered on the oscillometric pulse peaks that were then extracted.
A spectrogram was obtained from each beat, and all beats between the manually determined SBPs and DBPs were labeled as Korotkoff.
A three \textit{layer} CNN was then used to analyze consistency in sound patterns that were associated with Korotkoff sounds.
According to the authors this was the first study performed for this task providing evidence that it is difficult to identify Korotkoff sounds at systole and diastole.

\subsubsection{Data from wearable devices}
Wearable devices, which impose restrictions on size, power and memory consumption for models, have also been used to collect cardiology data for training deep learning models for AF detection.

Shashikumar et al.\cite{shashikumar2017deep} captured ECG, Pulsatile Photoplethysmographic (PPG) and accelerometry data from 98 subjects using a wrist-worn device and derived the spectrogram using continuous WT\@.
They trained a five \textit{layer} CNN in a sequence of short windows with movement artifacts and its output was combined with features calculated based on beat-to-beat variability and the signal quality index.
An accuracy of 91.8\% in AF detection was achieved by the method and in combination with its computational efficiency it is promising for real world deployment.
Gotlibovych et al.\cite{gotlibovych2018end} trained an one \textit{layer} CNN network followed by a LSTM using 180h of PPG wearable data to detect AF\@.
Use of the LSTM layer allows the network to learn variable-length correlations in contrast with the fixed length of the convolutional layer.
Poh et al.\cite{poh2018diagnostic} created a large database of PPG (over 180000 signals from 3373 patients) including data from MIMIC to classify four rhythms: sinus, noise, ectopic and AF\@.
A densely connected CNN with six blocks and a growth rate of six was used for classification that was fed with 17 second segments that have been denoised using a bandpass filter.
Results were obtained using an independent dataset of 3039 PPG achieving better results than previous methods that were based on handcrafted features.

Besides AF detection, wearable data have been used to search for optimal cardiovascular disease predictors.
In\cite{ballinger2018deepheart} the authors trained a semi-supervised, multi-task bi-directional LSTM on data from 14011 users of the Cardiogram app for detecting diabetes, high cholesterol, high BP, and sleep apnoea.
Their results indicate that the heart's response to physical activity is a salient biomarker for predicting the onset of a disease and can be captured using deep learning.

\section{Deep learning using imaging modalities}
\label{sec:imaging}
Imaging modalities that have found use in cardiology include Magnetic Resonance Imaging (MRI), Fundus Photography, Computerized Tomography (CT), Echocardiography, Optical Coherence Tomography (OCT), Intravascular Ultrasound (IVUS), and others.
Deep learning has been mostly successful in this area, mainly due to architectures that make use of convolutional layers and network depth.
A summary of deep learning applications using images are shown in Tables~\ref{table:imaging1},~\ref{table:imaging2}, and~\ref{table:imaging3}.

\subsection{Magnetic resonance imaging}
MRI is based on the interaction between a system of atomic nuclei and an external magnetic field providing a picture of the interior of a physical object\cite{sebastiani1991mathematical}.
The main uses of MRI include Left Ventricle (LV), Right Ventricle (RV) and whole heart segmentation.

\begin{table*}[!t]
	\caption{Deep learning applications using MRI}
	\label{table:imaging1}
	\begin{minipage}{\textwidth}
		\centering
		\begin{tabularx}{\textwidth}{l c l l}
			\toprule
			\thead{Reference}                          & \thead{Method} & \thead{Application/Notes\footnote{In parenthesis the databases used.}}                    & \thead{Dice\footnote{($\S$) denotes for `for each database', ($*$) denotes mean square error for EF ($+$) denotes `for endocardial and epicardial', ($-$) denotes accuracy, ($\#$) denotes `for CT and MRI'}} \\
			\midrule
			\multicolumn{4}{l}{\thead{LV segmentation}}                                                                                                                                                                                                                                                                                                                                                              \\
			\midrule
			Tan 2016\cite{tan2016cardiac}              & CNN            & CNN for localization and CNN for delineation of endocardial border (SUN09, STA11)         & 88\%                                                                                                                                                                                                                                           \\
			Romaguera 2017\cite{romaguera2017left}     & CNN            & five \textit{layer} CNN with SGD and RMSprop (SUN09)                                      & 92\%                                                                                                                                                                                                                                           \\
			Poudel 2016\cite{poudel2016recurrent}      & u-net, RNN     & combine u-net and RNN (SUN09, private)                                                    & \textit{multiple}                                                                                                                                                                                                                                     \\
			Rupprecht 2016\cite{rupprecht2016deep}     & CNN            & combine four \textit{layer} CNN with Sobolev (STA11, non-medical)                         & 85\%                                                                                                                                                                                                                                           \\
			Ngo 2014\cite{anh2014fully}                & DBN            & combine DBN with level set (SUN09)                                                        & 88\%                                                                                                                                                                                                                                           \\
			Avendi 2016\cite{avendi2016combined}       & CNN, AE        & CNN for chamber detection, AEs for shape inference and deformable models (SUN09)          & 96.69\%                                                                                                                                                                                                                                        \\
			Yang 2016\cite{yang2016deep}               & CNN            & feature extraction network and a non-local patch-based label fusion network (SAT13)       & 81.6\%                                                                                                                                                                                                                                         \\
			Luo 2016\cite{luo2016cardiac}              & CNN            & a LV atlas mapping method and a three \textit{layer} CNN (DS16)                           & 4.98\%$^*$                                                                                                                                                                                                                                     \\
			Yang 2017\cite{yang2017deep}               & CNN, u-net     & localization with regression CNN and segmentation with u-net (YUDB, SUN09)                & 91\%, 93\%$^\S$                                                                                                                                                                                                                                     \\
			Tan 2017\cite{tan2017convolutional}        & CNN            & regression CNN (STA11, DS16)                                                              & \textit{multiple}                                                                                                                                                                                                                                       \\
			Curiale 2017\cite{curiale2017automatic}    & u-net          & use residual u-net (SUN09)                                                                & 90\%                                                                                                                                                                                                                                           \\
			Liao 2017\cite{liao2017estimation}         & CNN            & local binary pattern for localization and hypercolumns FCN for segmentation (DS16)        & 4.69\%$^*$                                                                                                                                                                                                                                     \\
			Emad 2015\cite{emad2015automatic}          & CNN            & LV localization using CNN and pyramid of scales (YUDB)                                    & 98.66\%$^-$                                                                                                                                                                                                                                    \\
			\midrule
			\multicolumn{4}{l}{\thead{LV/RV segmentation}}                                                                                                                                                                                                                                                                                                                                                           \\
			\midrule
			Zotti 2017\cite{zotti2017gridnet}          & u-net          & u-net variant with a multi-resolution conv-deconv grid architecture (AC17)                & 90\%                                                                                                                                                                                                                                           \\
			Patravali 2017\cite{patravali20172d}       & u-net          & 2D/3D u-net trained (AC17)                                                                & \textit{multiple}                                                                                                                                                                                                                                       \\
			Isensee 2017\cite{isensee2017automatic}    & u-net          & ensemble of u-net, regularized multi-layer perceptrons and a RF classifier (AC17)         & \textit{multiple}                                                                                                                                                                                                                                       \\
			Tran 2016\cite{tran2016fully}              & CNN            & four \textit{layer} FCN (SUN09, STA11)                                                    & 92\%, 96\%$^+$                                                                                                                                                                                                                                 \\
			Bai 2017\cite{bai2017semi}                 & CNN            & VGG-16 and DeepLab architecture with use of CRF for refined results (UKBDB)               & 90.3\%                                                                                                                                                                                                                                         \\
			Lieman 2017\cite{lieman2017fastventricle}  & u-net          & extension of ENet\cite{paszke2016enet} with skip connections (private)                    & \textit{multiple}                                                                                                                                                                                                                                       \\
			Winther 2017\cite{winther2017nu}           & u-net          & $\nu$-net is a u-net variant (DS16, SUN09, RV12, private)                                 & \textit{multiple}                                                                                                                                                                                                                                       \\
			Du 2018\cite{du2018deep}                   & DBN            & DAISY features and regression DBN using 2900 images (private)                             & 91.6\%, 94.1\%$^+$                                                                                                                                                                                                                             \\
			Giannakidis 2016\cite{giannakidis2016fast} & CNN            & RV segmentation using 3D multi-scale CNN with two pathways (private)                      & 82.81\%                                                                                                                                                                                                                                        \\
			\midrule
			\multicolumn{4}{l}{\thead{Whole heart segmentation}}                                                                                                                                                                                                                                                                                                                                                     \\
			\midrule
			Wolterink 2016\cite{wolterink2016dilated}  & CNN            & dilated CNN with orthogonal patches (HVS16)                                               & 80\%, 93\%                                                                                                                                                                                                                                     \\
			Li 2016\cite{li2016automatic}              & CNN            & deeply supervised 3D FCN with dilations (HVS16)                                           & 69.5\%                                                                                                                                                                                                                                         \\
			Yu 2017\cite{yu20163d}                     & CNN            & deeply supervised 3D FCN constructed in a self-similar fractal scheme (HVS16)             & \textit{multiple}                                                                                                                                                                                                                                       \\
			Payer\cite{payer2017multi}                 & CNN            & FCN for localization and another FCN for segmentation (MM17)                              & 90.7\%, 87\%$^\#$                                                                                                                                                                                                                                     \\
			Mortazi 2017\cite{mortazi2017multi}        & CNN            & multi-planar FCN (MM17)                                                                   & 90\%, 85\%$^\#$                                                                                                                                                                                                                                     \\
			Yang\cite{yang2017hybrid}                  & CNN            & deeply supervised 3D FCN trained with transfer learning (MM17)                            & 84.3\%, 77.8\%$^\#$                                                                                                                                                                                                                                 \\
			\midrule
			\multicolumn{4}{l}{\thead{Others applications}}                                                                                                                                                                                                                                                                                                                                                          \\
			\midrule
			Yang 2017\cite{yang2017segmenting}         & SSAE           & atrial fibrosis segmentation using multi-atlas propagation, SSAE and softmax (private)    & 82\%                                                                                                                                                                                                                                           \\
			Zhang 2016\cite{zhang2016automated}        & CNN            & missing apical and basal identification using two CNNs with four \textit{layers} (UKBDB)  & \textit{multiple}                                                                                                                                                                                                                                       \\
			Kong 2016\cite{kong2016recognizing}        & CNN, RNN       & CNN for spatial information and RNN for temporal information to identify frames (private) & \textit{multiple}                                                                                                                                                                                                                                  \\
			Yang 2017\cite{yang2017convolutional}      & CNN            & CNN to identify end-diastole and end-systole frames from LV (STA11, private)              & 76.5\%$^-$                                                                                                                                                                                                                                     \\
			Xu 2017\cite{xu2017direct}                 & Multiple       & MI detection using Fast R-CNN for heart localization, LSTM and SAE (private)              & 94.3\%$^-$                                                                                                                                                                                                                                     \\
			Xue 2018\cite{xue2018full}                 & CNN, LSTM      & CNN, two parallel LSTMs and a Bayesian framework for full LV quantification (private)     & \textit{multiple}                                                                                                                                                                                                                                       \\
			Zhen 2016\cite{zhen2016multi}              & RBM            & multi-scale convolutional RBM and RF for bi-ventricular volume estimation (private)       & 3.87\%$^*$                                                                                                                                                                                                                                     \\
			Biffi 2016\cite{biffi2018learning}         & CNN            & identify hypertrophic cardiomyopathy using a variational AE (AC17, private)               & 90\%$^-$                                                                                                                                                                                                                                       \\
			Oktay 2016\cite{oktay2016multi}            & CNN            & image super resolution using residual CNN (private)                                       & \textit{multiple}                                                                                                                                                                                                                                       \\
			\bottomrule
		\end{tabularx}
	\end{minipage}
\end{table*}

\subsubsection{Left ventricle segmentation}
CNNs were used for LV segmentation with MRI\@.
Tan et al.\cite{tan2016cardiac} used a CNN to localize LV endocardium and another CNN to determine the endocardial radius using STA11 and SUN09 for training and evaluation respectively.
Without filtering out apical slices and using deformable models they achieve comparable performance with previous state-of-the-art methods.
In\cite{romaguera2017left} the authors trained a five \textit{layer} CNN using MRI from SUN09 challenge.
They trained their model using SGD and RMSprop with the former achieving a better Dice of 92\%.

CNNs combined with RNNs were also used.
In\cite{poudel2016recurrent} the authors created a recurrent u-net that learns image representations from a stack of 2D slices and has the ability to leverage inter-slice spatial dependencies through internal memory units.
It combines anatomical detection and segmentation into a single end-to-end architecture, achieving comparable results with other non end-to-end methods, outperforming the baselines DBN, recurrent DBN and FCN in terms of Dice.

Other papers combined deep learning methods with level set for LV segmentation.
Rupprecht et al.\cite{rupprecht2016deep} trained a class-specific four \textit{layer} CNN which predicts a vector pointing from the respective point on the evolving contour towards the closest point on the boundary of the object of interest.
These predictions formed a vector field which was then used for evolving the contour using the Sobolev active contour framework.
Anh et al.\cite{anh2014fully} created a non-rigid segmentation method based on the distance regularized level set method that was initialized and constrained by the results of a structured inference using a DBN\@.
Avendi et al.\cite{avendi2016combined} used CNN to detect the LV chamber and then utilized stacked AEs to infer the shape of the LV\@.
The result was then incorporated into deformable models to improve the accuracy and robustness of the segmentation.

Atlas-based methods have also been used for this task.
Yang et al.\cite{yang2016deep} created an end-to-end deep fusion network by concatenating a feature extraction network and a non-local patch-based label fusion network.
The learned features are further utilized in defining a similarity measure for MRI atlas selection.
They compared their method with majority voting, patch-based label fusion, multi-atlas patch match and SVM with augmented features achieving superior results in terms of accuracy.
Luo et al.\cite{luo2016cardiac} adopted a LV atlas mapping method to achieve accurate localization using MRI data from DS16.
Then, a three \textit{layer} CNN was trained for predicting the LV volume, achieving comparable results with the winners of the challenge in terms of root mean square of end-diastole and end-systole volumes.

Regression-based methods have been used for localizing the LV before segmenting it.
Yang et al.\cite{yang2017deep} first locate LV in the full image using a regression CNN and then segment it within the cropped region of interest using a u-net based architecture.
They demonstrate that their model achieves high accuracy with computational performance during inference.

Various other methods were used.
Tan et al.\cite{tan2017convolutional} parameterize all short axis slices and phases of the LV segmentation task in terms of the radial distances between the LV center-point and the endocardial and epicardial contours in polar space.
Then, they train a CNN regression on STA11 to infer these parameters and test the generalizability of the method on DS16 with good results.
In\cite{curiale2017automatic} the authors used Jaccard distance as optimization objective function, integrating a residual learning strategy, and introducing a batch normalization layer to train a u-net.
It is shown in the paper that this configuration performed better than other simpler u-nets in terms of Dice.
In their article Liao et al.\cite{liao2017estimation} detected the Region of Interest (ROI) containing LV chambers and then used hypercolumns FCN to segment LV in the ROI\@.
The 2-D segmentation results were integrated across different images to estimate the volume.
The model was trained alternately on LV segmentation and volume estimation, placing fourth in the test set of DS16.
Emad et al.\cite{emad2015automatic} localize the LV using a CNN and a pyramid of scales analysis to take into account different sizes of the heart with the YUDB\@.
They achieve good results but with a significant computation cost (10 seconds per image during inference).

\subsubsection{LV/RV segmentation}
A dataset used for LV/RV segmentation was the MICCAI 2017 ACDC Challenge (AC17) that contains MRI images from 150 patients divided into five groups (normal, previous MI, dilated cardiomyopathy, hypertrophic cardiomyopathy, abnormal RV).
Zotti et al.\cite{zotti2017gridnet} used a model that includes a cardiac center-of-mass regression module which allows shape prior registration and a loss function tailored to the cardiac anatomy.
Features are learned with a multi-resolution conv-deconv `grid' architecture which is an extension of u-net.
This model compared with vanilla conv-deconv and u-net performs better by an average of 5\% in terms of Dice.
Patravali et al.\cite{patravali20172d} trained a model based on u-net using Dice combined with cross entropy as a metric for LV/RV and myocardium segmentation.
The model was designed to accept a stack of image slices as input channels and the output is predicted for the middle slice.
Based on experiments they conducted, it was concluded that three input slices were optimal as an input for the model, instead of one or five.
Isensee et al.\cite{isensee2017automatic} used an ensemble of a 2D and a 3D u-net for segmentation of the LV/RV cavity and the LV myocardium on each time instance of the cardiac cycle.
Information was extracted from the segmented time-series in form of features that reflect diagnostic clinical procedures for the purposes of the classification task.
Based on these features they then train an ensemble of regularized multi-layer perceptrons and a RF classifier to predict the pathological target class.
Their model ranked first in the ACDC challenge.

Various other datasets have also been used for this task with CNNs.
In\cite{bai2017semi} the authors created a semi-supervised learning method, in which a segmentation network for LV/RV and myocardium was trained from labeled and unlabeled data.
The network architecture was adapted from VGG-16, similar to the DeepLab architecture\cite{chen2018deeplab} while the final segmentation was refined using a Conditional Random Field (CRF)\@.
The authors show that the introduction of unlabelled data improves segmentation performance when the training set is small.
In\cite{giannakidis2016fast} the authors adopt a 3D multi-scale CNN to identify pixels that belong to the RV\@.
The network has two convolutional pathways and their inputs are centered at the same image location, but the second segment is extracted from a down-sampled version of the image.
The results obtained were better than the previous state-of-the-art although the latter were based on feature engineering and trained on less variable datasets.

FCNs have also been used for LV/RV segmentation.
In their article Tran et al.\cite{tran2016fully} trained a four \textit{layer} FCN model for LV/RV segmentation on SUN09, STA11.
They compared previous state-of-the-art methods along with two initializations of their model: a fine-tuned version of their model using STA11 and a Xavier initialized model with the former performing best in almost all tasks.

FCNs with skip connections and u-net have also been used for this task.
Lieman et al.\cite{lieman2017fastventricle} created a FCN architecture with skip connections named FastVentricle based on ENet\cite{paszke2016enet} which is faster and runs with less memory than previous ventricular segmentation architectures achieving high clinical accuracy.
In\cite{winther2017nu} the authors introduce $\nu$-net which is a u-net variant for segmentation of LV/RV endocardium and epicardium using DS16, SUN09 and RV12 datasets.
This method performed better than the expert cardiologist in this study, especially for RV segmentation.

Some methods were based on regression models.
In their article Du et al.\cite{du2018deep} created a regression segmentation framework to delineate boundaries of LV/RV\@.
First, DAISY features are extracted and then a point-based representation method was employed to depict the boundaries.
Finally, the DAISY features were used as input and the boundary points as labels to train the regression model based on DBN\@.
The model performance is evaluated using different features than DAISY (GIST, pyramid histogram of oriented gradients) and also compared with support vector regression and other traditional methods (graph cuts, active contours, level set) achieving better results.

\subsubsection{Whole heart segmentation}
MICCAI 2016 HVSMR (HVS16) was used for whole heart segmentation that contains MRI images from 20 patients.
Wolterink et al.\cite{wolterink2016dilated} trained a ten \textit{layer} CNN with increasing levels of dilation for segmenting the myocardium and blood pool in axial, sagittal and coronal image slices.
They also employ deep supervision\cite{lee2015deeply} to alleviate the vanishing gradients problem and improve the training efficiency of their network using a small dataset.
Experiments performed with and without dilations on this architecture indicated the usefulness of this configuration.
In their article Li et al.\cite{li2016automatic} start with a 3D FCN for voxel-wise labeling and then introduce dilated convolutional layers into the baseline model to expand its receptive field.
Then, they employ deep-supervised pathways to accelerate training and exploit multi-scale information.
According to the authors the model demonstrates good segmentation accuracy combined with low computational cost.
Yu et al.\cite{yu20163d} created a 3D FCN fractal network for whole heart and great vessel volume-to-volume segmentation.
By recursively applying a single expansion rule, they construct the network in a self-similar fractal scheme combining hierarchical clues for accurate segmentation.
They also achieve good results with low computational cost (12 seconds per volume).

Another database used for whole heart segmentation was MM17 which contains 120 multimodal images from cardiac MRI/CT\@.
Payer et al.\cite{payer2017multi} method is based on two FCN for multi-label whole heart localization and segmentation.
At first, the localization CNN finds the center of the bounding box around all heart structures, such that the segmentation CNN can focus on this region.
Trained in an end-to-end manner, the segmentation CNN transforms intermediate label predictions to positions of other labels.
Therefore, the network learns from the relative positions among labels and focuses on anatomically feasible configurations.
The model was compared with u-net achieving superior results, especially in the MRI dataset.
Mortazi et al.\cite{mortazi2017multi} trained a multi-planar CNN with an adaptive fusion strategy for segmenting seven substructures of the heart.
They designed three CNNs, one for each plane, with the same architectural configuration and trained them for voxel-wise labeling.
Their experiments conclude that their model delineates cardiac structures with high accuracy and efficiency.
In\cite{yang2017hybrid} the authors used FCN, and couple it with 3D operators, transfer learning and a deep supervision mechanism to distill 3D contextual information and solve potential difficulties in training.
A hybrid loss was used that guides the training procedure to balance classes, and preserve boundary details.
According to their experiments, using the hybrid loss achieves better results than using only Dice.

\subsubsection{Other tasks}
Deep learning papers has also been used for detection of other cardiac structures with MRI\@.
Yang et al.\cite{yang2017segmenting} created a multi-atlas propagation method to derive the anatomical structure of the left atrium myocardium and pulmonary veins.
This was followed by a unsupervised trained SSAE with a softmax for atrial fibrosis segmentation using 20 scans from AF patients.
In their article Zhang et al.\cite{zhang2016automated} try to detect missing apical and basal slices.
They test the presence of typical basal and apical patterns at the bottom and top slices of the dataset and train two CNNs to construct a set of discriminative features.
Their experiments showed that the model with four \textit{layers} performed better than the baselines SAE and Deep Boltzmann Machines.

Other medical tasks in MRI were also studied such as detection of end-diastole end-systole frames.
Kong et al.\cite{kong2016recognizing} created a temporal regression network pretrained on ImageNet by integrating a CNN with a RNN, to identify end-diastole and end-systole frames from MRI sequences.
The CNN encodes the spatial information of a cardiac sequence, and the RNN decodes the temporal information.
They also designed a loss function to constrain the structure of predicted labels.
The model achieves better average frame difference than the previous methods.
In their article Yang et al.\cite{yang2017convolutional} used a CNN to identify end-diastole and end-systole frames from the LV, achieving an overall accuracy of 76.5\%.

There were also methods that tried to quantify various cardiovascular features.
In\cite{xu2017direct} the authors detect the area, position and shape of the MI using a model that consists of three \textit{layers}; first, the heart localization layer is a Fast R-CNN\cite{girshick2015fast} which crops the ROI sequences including the LV; second, the motion statistical layers, which build a time-series architecture to capture the local motion features generated by LSTM-RNN and the global motion features generated by deep optical flows from the ROI sequence; third, the fully connected discriminate layers, which use SAE to further learn the features from the previous layer and a softmax classifier.
Xue et al.\cite{xue2018full} trained an end-to-end deep multitask relationship learning framework on MRI images from 145 subjects with 20 frames each for full LV quantification.
It consists of a three \textit{layer} CNN that extracts cardiac representations, then two parallel LSTM-based RNNs for modeling the temporal dynamics of cardiac sequences.
Finally, there is a Bayesian framework capable of learning multitask relationships and a softmax classifier for classification.
Extensive comparisons with the state-of-the-art show the effectiveness of this method in terms of mean absolute error.
In\cite{zhen2016multi} the authors created an unsupervised cardiac image representation learning method using multi-scale convolutional RBM and a direct bi-ventricular volume estimation using RF\@.
They compared their model with a Bayesian model, a feature based model, level sets and graph cut achieving better results in terms of correlation coefficient for LV/RV volumes and estimation error of EF\@.

Other methods were also created to detect hypertrophic cardiomyopathy or increase the resolution of MRI\@.
Biffi et al.\cite{biffi2018learning} trained a variational AE to identify hypertrophic cardiomyopathy subjects tested on a multi-center balanced dataset of 1365 patients and AC17.
They also demonstrate that the network is able to visualize and quantify the learned pathology-specific remodeling patterns in the original input space of the images, thus increasing the interpretability of the model.
In\cite{oktay2016multi} the authors created an image super-resolution method based on a residual CNN that allows the use of input data acquired from different viewing planes for improved performance.
They compared it with other interpolation methods (linear, spline, multi-atlas patch match, shallow CNN, CNN) achieving better results in terms of PSNR\@.
Authors from the same group proposed a training strategy\cite{oktay2018anatomically} that incorporates anatomical prior knowledge into CNNs through a regularization model, by encouraging it to follow the anatomy via learned non-linear representations of the shape.

\subsubsection{Overall view on deep learning using MRI}
There is a wide range of architectures that have been applied in MRI\@.
Most predominantly CNNs and u-nets are used solely or in combination with RNNs, AEs, or ensembles.
The problem is that most of them are not end-to-end; they rely on preprocessing, handcrafted features, active contours, level set and other non-differentiable methods, thus partially losing the ability to scale on the presence of new data.
The main target of this area should be to create end-to-end models even if that means less accuracy in the short-term; more efficient architectures could close the gap in the future.

An interesting finding regarding whole heart segmentation was done in\cite{konukoglu2018exploration} where the authors investigated the suitability of state-of-the-art 2D, 3D CNN architectures, and modifications of them.
They find that processing the images in a slice-by-slice fashion using 2D networks was beneficial due to the large slice thickness.
However, the choice of the network architecture plays a minor role.

\subsection{Fundus photography}
Fundus photography is a clinical tool for evaluating retinopathy progress in patients where the image intensity represents the amount of reflected light of a specific waveband\cite{abramoff2010retinal}.
One of the most widely used databases in fundus is DRIVE which contains 40 images and their corresponding vessel mask annotations.

\begin{table*}[!t]
	\caption{Deep learning applications using fundus photography}
	\label{table:imaging2}
	\begin{minipage}{\textwidth}
		\centering
		\begin{tabularx}{\textwidth}{l c l l}
			\toprule
			\thead{Reference}                                & \thead{Method} & \thead{Application/Notes\footnote{In parenthesis the databases used.}}                             & \thead{AUC\footnote{($*$) denotes accuracy.}} \\
			\midrule
			\multicolumn{4}{l}{\thead{Vessel segmentation}}                                                                                                                                                                        \\
			\midrule
			Wang 2015\cite{wang2015hierarchical}             & CNN, RF        & three \textit{layer} CNN combined with ensemble RF (DRIVE, STARE)                                  & 0.9475                                       \\
			Zhou 2017\cite{zhou2017improving}                & CNN, CRF       & CNN to extract features and CRF for final result (DRIVE, STARE, CHDB)                              & 0.7942                                       \\
			Chen 2017\cite{chen2017labeling}                 & CNN            & artificial data, FCN (DRIVE, STARE)                                                                & 0.9516                                       \\
			Maji 2016\cite{maji2016ensemble}                 & CNN            & 12 CNNs ensemble with three \textit{layers} (DRIVE)                                                & 0.9283                                       \\
			Fu 2016\cite{fu2016retinal}                      & CNN, CRF       & CNN and CRF (DRIVE, STARE)                                                                         & 94.70\%$^*$                                   \\
			Wu 2016\cite{wu2016deep}                         & CNN            & vessel segmentation and branch detection using CNN and PCA (DRIVE)                                 & 0.9701                                       \\
			Li 2016\cite{li2016cross}                        & SDAE           & FNN and SDAE (DRIVE, STARE, CHDB)                                                                  & 0.9738                                       \\
			Lahiri 2016\cite{lahiri2016deep}                 & SDAE           & ensemble of two level of sparsely trained SDAE (DRIVE)                                             & 95.30\%$^*$                                   \\
			Oliveira 2017\cite{oliveira2017augmenting}       & u-net          & data augmentation and u-net (DRIVE)                                                                & 0.9768                                       \\
			Leopold 2017\cite{leopold2017use}                & CNN            & CNN as a multi-channel classifier and Gabor filters (DRIVE)                                        & 94.78\%$^*$                                   \\
			Leopold 2017\cite{leopold2017pixelbnn}           & AE             & fully residual AE with gated streams based on u-net (DRIVE, STARE, CHDB)                           & 0.8268                                       \\
			Mo 2017\cite{mo2017multi}                        & CNN            & auxiliary classifiers and transfer learning (DRIVE, STARE, CHDB)                                   & 0.9782                                       \\
			Melinscak 2015\cite{melinvsvcak2015retinal}      & CNN            & four \textit{layer} CNN (DRIVE)                                                                    & 0.9749                                       \\
			Sengur 2017\cite{sengur2017retinal}              & CNN            & two \textit{layer} CNN with dropout (DRIVE)                                                        & 0.9674                                       \\
			Meyer 2017\cite{meyer2017deep}                   & u-net          & vessel segmentation using u-net on SLO (IOSTAR, RC-SLO)                                            & 0.9771                                       \\
			\midrule
			\multicolumn{4}{l}{\thead{Microaneurysm and hemorrhage detection}}                                                                                                                                                     \\
			\midrule
			Haloi 2015\cite{haloi2015improved}               & CNN            & MA detection using CNN with dropout and maxout activation (ROC, Messidor, DIA)                     & 0.98                                          \\
			Giancardo 2017\cite{giancardo2017representation} & u-net          & MA detection using internal representation of trained u-net (DRIVE, Messidor)                      & \textit{multiple}                                      \\
			Orlando 2018\cite{orlando2018ensemble}           & CNN            & MA and hemorrhage detection using handcrafted features and a CNN (DIA, e-optha, Messidor)          & \textit{multiple}                                      \\
			van Grinsven 2017\cite{van2016fast}              & CNN            & hemorrhage detection with selective data sampling using a five \textit{layer} CNN (KR15, Messidor) & \textit{multiple}                                      \\
			\midrule
			\multicolumn{4}{l}{\thead{Other applications}}                                                                                                                                                                         \\
			\midrule
			Girard 2017\cite{girard2017artery}               & CNN            & artery/vein classification using CNN and likelihood score propagation (DRIVE, Messidor)            & \textit{multiple}                                      \\
			Welikala 2017\cite{welikala2017automated}        & CNN            & artery/vein classification using three \textit{layer} CNN (UKBDB)                                  & 82.26\%$^*$                                   \\
			Pratt 2017\cite{pratt2017automatica}             & ResNet         & bifurcation/crossing classification using ResNet 18 (DRIVE, IOSTAR)                                & \textit{multiple}                                      \\
			Poplin 2017\cite{poplin2017predicting}           & Inception      & cardiovascular risk factors prediction (UKBDB, private)                                            & \textit{multiple}                                      \\
			\bottomrule
		\end{tabularx}
	\end{minipage}
\end{table*}

\subsubsection{Vessel segmentation}
CNNs have been used for vessel segmentation in fundus imaging.
In\cite{wang2015hierarchical} the authors first used histogram equalization and Gaussian filtering to reduce noise.
A three \textit{layer} CNN was then used as a feature extractor and a RF as the classifier.
According to experiments done by the authors the best performance was achieved by a winner-takes-all ensemble, compared with an average, weighted and median ensemble.
Zhou et al.\cite{zhou2017improving} applied image preprocessing to eliminate the strong edges around the field of view and normalize the luminosity and contrast inside it.
Then, they trained a CNN to generate features for linear models and applied filters to enhance thin vessels, reducing the intensity difference between thin and wide vessels.
A dense CRF was then adapted to achieve the final retinal vessel segmentation, by taking the discriminative features for unary potentials and the thin-vessel enhanced image for pairwise potentials.
Amongst their results, in which they demonstrate better accuracy than most state-of-the-art methods, they also provide evidence in favor of using the RGB information of the fundus instead of just the green channel.
Chen\cite{chen2017labeling} designed a set of rules to generate artificial training samples with prior knowledge and without manual labeling.
They train a FCN with a concatenation layer that allows high level perception guide the work in lower levels and evaluate their model on DRIVE and STARE databases, achieving comparable results with other methods that use real labeling.
In\cite{maji2016ensemble} the authors trained a 12 CNNs ensemble with three \textit{layers} each on the DRIVE database, where during inference the responses of the CNNs are averaged to form the final segmentation.
They demonstrate that their ensemble achieves higher maximum average accuracy than previous methods.
Fu et al.\cite{fu2016retinal} train a CNN on DRIVE and STARE databases to generate the vessel probability maps and then they employed a fully connected CRF to combine the discriminative vessel probability maps and long-range interactions between pixels.
In\cite{wu2016deep} the authors used a CNN to learn the features and a PCA-based nearest neighbor search utilized to estimate the local structure distribution.
Besides demonstrating good results they argue that it is important for CNN to incorporate information regarding the tree structure in terms of accuracy.

AEs were used for vessel segmentation.
Li et al.\cite{li2016cross} trained a FNN and a denoising AE with DRIVE, STARE and CHDB databases.
They argue that the learnt features of their model are more reliable to pathology, noise and different imaging conditions, because the learning process exploits the characteristics of vessels in all training images.
In\cite{lahiri2016deep} the authors employed unsupervised hierarchical feature learning using a two level ensemble of sparsely trained SDAE\@.
The training level ensures decoupling and the ensemble level ensures architectural revision.
They show that ensemble training of AEs fosters diversity in learning dictionary of visual kernels for vessel segmentation.
Softmax classifier was then used for fine-tuning each AE and strategies are explored for two level fusion of ensemble members.

Other architectures were also used for vessel segmentation.
In their article Oliveira et al.\cite{oliveira2017augmenting} trained a u-net with DRIVE demonstrating good results and presenting evidence of the benefits of data augmentation on the training data using elastic transformations.
Leopold et al.\cite{leopold2017use} investigated the use of a CNN as a multi-channel classifier and explore the use of Gabor filters to boost the accuracy of the method described in\cite{leopold2017segmentation}.
They applied the mean of a series of Gabor filters with varying frequencies and sigma values to the output of the network to determine whether a pixel represents a vessel or not.
Besides finding that the optimal filters vary between channels, the authors also state the `need' of enforcing the networks to align with human perception, in the context of manual labeling, even if that requires downsampling information, which would otherwise reduce the computational cost.
The same authors\cite{leopold2017pixelbnn} created PixelBNN which is a fully residual AE with gated streams.
It is more than eight times faster than the previous state-of-the-art methods at test time and performed well, considering a significant reduction in information from resizing images during preprocessing.
In their article Mo et al.\cite{mo2017multi} used deep supervision with auxiliary classifiers in intermediate layers of the network, to improve the discriminative capability of features in lower layers of the deep network and guide backpropagation to overcome vanishing gradient.
Moreover, transfer learning was used to overcome the issue of insufficient medical training data.

\subsubsection{Microaneurysm and hemorrhage detection}
Haloi\cite{haloi2015improved} trained a three \textit{layer} CNN with dropout and maxout activation function for MA detection.
Experiments on ROC and DIA demonstrated state-of-the-art results.
In\cite{giancardo2017representation} the authors created a model that learns a general descriptor of the vasculature morphology using the internal representation of a u-net variation.
Then, they tested the vasculature embeddings on a similar image retrieval task according to vasculature and on a diabetic retinopathy classification task, where they show how the vasculature embeddings improve the classification of a method based on MA detection.
In\cite{orlando2018ensemble} the authors combined augmented features learned by a CNN with handcrafted features.
This ensemble vector of descriptors was then used to identify MA and hemorrhage candidates using a RF classifier.
Their analysis using t-SNE demonstrates that CNN features have fine-grained characteristics such as the orientation of the lesion while handcrafted features are able to discriminate low contrast lesions such as hemorrhages.
In\cite{van2016fast} the authors trained a five \textit{layer} CNN to detect hemorrhage using 6679 images from DS16 and Messidor databases.
They applied selective data sampling on a CNN which increased the speed of the training by dynamically selecting misclassified negative samples during training.
Weights are assigned to the training samples and informative samples are included in the next training iteration.

\subsubsection{Other tasks}
Fundus has also been used for artery/vein classification.
In their article Girard et al.\cite{girard2017artery} trained a four \textit{layer} CNN that classifies vessel pixels into arteries/veins using rotational data augmentation.
A graph was then constructed from the retinal vascular network where the nodes are defined as the vessel branches and each edge gets associated to a cost that evaluates whether the two branches should have the same label.
The CNN classification was propagated through the minimum spanning tree of the graph.
Experiments demonstrated the effectiveness of the method especially in the presence of occlusions.
Welikala et al.\cite{welikala2017automated} trained and evaluated a three \textit{layer} CNN using centerline pixels derived from retinal images.
Amongst their experiments they found that rotational and scaling data augmentations did not help increase accuracy, attributing it to interpolation altering pixel intensities which is problematic due to the sensitivity of CNN to pixel distribution patterns.

There are also other uses such as bifurcation/crossing identification.
Pratt et al.\cite{pratt2017automatica} trained a ResNet18 to identify small patches which include either bifurcation or crossing.
Another ResNet18 was trained on patches that have been classified to have bifurcations and crossings to distinguish the type of vessel junction located.
Similar work on this problem has been done by the same authors\cite{pratt2017automaticb} using a CNN\@.

An important result in the area of Cardiology using fundus photography is from Poplin et al.\cite{poplin2017predicting} who used an Inception v3 to predict cardiovascular risk factors (age, gender, smoking status, HbA1c, SBP) and major cardiac events.
Their models used distinct aspects of the anatomy to generate each prediction, such as the optic disc or the blood vessels, as it was demonstrated using the soft attention technique.
Most results were significantly better than previously thought possible with fundus photography (\textgreater{70\%} AUC).

\subsubsection{Overall view on deep learning using fundus}
Regarding the use of architectures there is a clear preference for CNNs especially in vessel segmentation while an interesting approach from some publications is the use of CRFs as a postprocessing step for vessel segmentation refinement\cite{zhou2017improving, fu2016retinal}.
The fact that there are many publicly available databases and that the DRIVE database is predominantly used in most of the literature makes this field easier to compare and validate new architectures.
Moreover the non-invasive nature of fundus and its recent use as a tool to estimate cardiovascular risk predictors makes it a promising modality of increased usefulness in the field of cardiology.

\subsection{Computerized tomography}
Computerized Tomography (CT) is a non-invasive method for the detection of obstructive artery disease.
Some of the areas that deep learning was applied with CT include coronary artery calcium score assessment, localization and segmentation of cardiac areas.

\begin{table*}[!t]
	\caption{Deep learning applications using CT, Echocardiography, OCT and other imaging modalities}
	\label{table:imaging3}
	\begin{minipage}{\textwidth}
		\centering
		\begin{tabularx}{\textwidth}{l c l}
			\toprule
			\thead{Reference}                            & \thead{Method}  & \thead{Application/Notes\footnote{Results from these imaging modalities are not reported in this review because they were highly variable in terms of the research question they were trying to solve and highly inconsistent in respect with the use of metrics. Additionally all papers use private databases besides\cite{liu2017left, tom2018simulating}.}} \\
			\midrule
			\multicolumn{3}{l}{\thead{CT}}                                                                                                                                                                                                                                                                    \\
			\midrule
			Lessman 2016\cite{lessmann2016deep}          & CNN             & detect coronary calcium using three independently trained CNNs                                                                                                                                                                   \\
			Shadmi 2018\cite{shadmi2018fully}            & DenseNet        & compared DenseNet and u-net for detecting coronary calcium                                                                                                                                                                       \\
			Cano 2018\cite{cano2018automated}            & CNN             & 3D regression CNN for calculation of the Agatston score                                                                                                                                                                          \\
			Wolterink 2016\cite{wolterink2016automatic}  & CNN             & detect coronary calcium using three CNNs for localization and two CNNs for detection                                                                                                                                             \\
			Santini 2017\cite{santini2017automatic}      & CNN             & coronary calcium detection using a seven \textit{layer} CNN on image patches                                                                                                                                                     \\
			Lopez 2017\cite{lopez2017dcnn}               & CNN             & thrombus volume characterization using a 2D CNN and postprocessing                                                                                                                                                               \\
			Hong 2016\cite{hong2016automatic}            & DBN             & detection, segmentation, classification of abdominal aortic aneurysm using DBN and image patches                                                                                                                                 \\
			Liu 2017\cite{liu2017left}                   & CNN             & left atrium segmentation using a twelve \textit{layer} CNN and active shape model (STA13)                                                                                                                                        \\
			de Vos 2016\cite{de20162d}                   & CNN             & 3D localization of anatomical structures using three CNNs, one for each orthogonal plane                                                                                                                                         \\
			Moradi 2016\cite{moradi2016hybrid}           & CNN             & detection of position for a given CT slice using a pretrained VGGnet, handcrafted features and SVM                                                                                                                               \\
			Zheng 2015\cite{zheng20153d}                 & Multiple        & carotid artery bifurcation detection using multi-layer perceptrons and probabilistic boosting-tree                                                                                                                               \\
			Montoya 2018\cite{montoya2018deep}           & ResNet          & 3D reconstruction of cerebral angiogram using a 30 layer ResNet                                                                                                                                                                  \\
			Zreik 2018\cite{zreik2018deep}               & CNN, AE         & identify coronary artery stenosis using CNN for LV segmentation and an AE, SVM for classification                                                                                                                                \\
			Commandeur 2018\cite{commandeur2018deep}     & CNN             & quantification of epicardial and thoracic adipose tissue from non-contrast CT                                                                                                                                                    \\
			Gulsun 2016\cite{gulsun2016coronary}         & CNN             & extract coronary centerline using optimal path from computed flow field and a CNN for refinement                                                                                                                                 \\
			\midrule
			\multicolumn{3}{l}{\thead{Echocardiography}}                                                                                                                                                                                                                                                      \\
			\midrule
			Carneiro 2012\cite{carneiro2012segmentation} & DBN             & LV segmentation by decoupling rigin and non-rigid detections using DBN on 480 images                                                                                                                                             \\
			Nascimento 2016\cite{nascimento2016multi}    & DBN             & LV segmentation using manifold learning and a DBN                                                                                                                                                                                \\
			Chen 2016\cite{chen2016iterative}            & CNN             & LV segmentation using multi-domain regularized FCN and transfer learning                                                                                                                                                         \\
			Madani 2018\cite{madani2018fast}             & CNN             & transthoracic echocardiogram view classification using a six \textit{layer} CNN                                                                                                                                                  \\
			Silva 2018\cite{silva2018ejection}           & CNN             & ejection fraction classification using a residual 3D CNN and transthoracic echocardiogram images                                                                                                                                 \\
			Gao 2017\cite{gao2017fused}                  & CNN             & viewpoint classification by fusing two CNNs with seven \textit{layers} each                                                                                                                                                      \\
			Abdi 2017\cite{abdi2017quality}              & CNN, LSTM       & assess quality score using convolutional and recurrent layers                                                                                                                                                                    \\
			Ghesu 2016\cite{ghesu2016marginal}           & CNN             & aortic valve segmentation using 2891 3D transesophageal echocardiogram images                                                                                                                                                    \\
			Perrin 2017\cite{perrin2017application}      & CNN             & congenital heart disease classification using a CNN trained in pairwise fashion                                                                                                                                                  \\
			Moradi 2016\cite{moradi2016cross}            & VGGnet, doc2vec & produce semantic descriptors for images                                                                                                                                                                                          \\
			\midrule
			\multicolumn{3}{l}{\thead{OCT}}                                                                                                                                                                                                                                                                   \\
			\midrule
			Roy 2016\cite{roy2016multiscale}             & AE              & tissue characterization using a distribution preserving AE                                                                                                                                                                       \\
			Yong 2017\cite{yong2017linear}               & CNN             & lumen segmentation using a linear-regression CNN with four \textit{layers}                                                                                                                                                       \\
			Xu 2017\cite{xu2017fibroatheroma}            & CNN             & presence of fibroatheroma using features extracted from previous architectures and SVM                                                                                                                                           \\
			Abdolmanafi 2017\cite{abdolmanafi2017deep}   & CNN             & intima, media segmentation using a pretrained AlexNet and comparing various classifiers                                                                                                                                          \\
			\midrule
			\multicolumn{3}{l}{\thead{Other imaging modalities}}                                                                                                                                                                                                                                              \\
			\midrule
			Lekadir 2017\cite{lekadir2017convolutional}  & CNN             & carotid plaque characterization using four \textit{layer} CNN on Ultrasound                                                                                                                                                      \\
			Tajbakhsh 2017\cite{tajbakhsh2017automatic}  & CNN             & carotid intima media thickness video interpretation using two CNNs with two \textit{layers} on Ultrasound                                                                                                                        \\
			Tom 2017\cite{tom2018simulating}             & GAN             & IVUS image generation using two GANs (IV11)                                                                                                                                                                                      \\
			Wang 2017\cite{wang2017detecting}            & CNN             & breast arterial calcification using a ten \textit{layer} CNN on mammograms                                                                                                                                                       \\
			Liu 2017\cite{liu2017coronary}               & CNN             & CAC detection using CNNs on 1768 X-Rays                                                                                                                                                                                          \\
			Pavoni 2017\cite{pavoni2017image}            & CNN             & denoising of percutaneous transluminal coronary angioplasty images using four \textit{layer} CNN                                                                                                                                 \\
			Nirschl 2018\cite{nirschl2018deep}           & CNN             & trained a patch-based six \textit{layer} CNN for identifying heart failure in endomyocardial biopsy images                                                                                                                       \\
			Betancur 2018\cite{betancur2018deep}         & CNN             & trained a three \textit{layer} CNN for obstructive CAD prediction from myocardial perfusion imaging                                                                                                                              \\
			\bottomrule
		\end{tabularx}
	\end{minipage}
\end{table*}

Deep learning was used for coronary calcium detection with CT.
Lessman et al.\cite{lessmann2016deep} method for coronary calcium scoring utilizes three independently trained CNNs to estimate a bounding box around the heart, in which connected components above a Hounsfield unit threshold are considered candidates for CACs.
Classification of extracted voxels was performed by feeding two-dimensional patches from three orthogonal planes into three concurrent CNNs to separate them from other high intensity lesions.
Patients were assigned to one of five standard cardiovascular risk categories based on the Agatston score.
Authors from the same group created a method\cite{lessmann2017automatic} for the detection of calcifications in low-dose chest CT using a CNN for anatomical location and another CNN for calcification detection.
In\cite{shadmi2018fully} the authors compared a four block u-net and a five block DenseNet for calculating the Agatston score using over 1000 images from a chest CT database.
The authors heavily preprocessed the images using thresholding, connected component analysis and morphological operations for lungs, trachea and carina detection.
Their experiments showed that DenseNet performed better in terms of accuracy.
Cano et al.\cite{cano2018automated} trained a three \textit{layer} 3D regression CNN that computes the Agatston score using 5973 non-ECG gated CT images achieving a Pearson correlation of 0.932.
In\cite{wolterink2016automatic} the authors created a method to identify and quantify CAC without a need for coronary artery extraction.
The bounding box detection around the heart method employs three CNNs, where each detects the heart in the axial, sagittal and coronal plane.
Another pair of CNNs were used to detect CAC\@.
The first CNN identifies CAC-like voxels, thereby discarding the majority of non-CAC-like voxels such as lung and fatty tissue.
The identified CAC-like voxels are further classified by the second CNN in the pair, which distinguishes between CAC and CAC-like negatives.
Although the CNNs share architecture, given that they have different tasks they do not share weights.
They achieve a Pearson correlation of 0.95, comparable with previous state-of-the-art.
Santini et al.\cite{santini2017automatic} trained a seven \textit{layer} CNN using patches for the segmentation and classification of coronary lesions in CT images.
They trained, validated and tested their network on 45, 18 and 56 CT volumes respectively achieving a Pearson correlation of 0.983.

CT has been used for segmentation of various cardiac areas.
Lopez et al.\cite{lopez2017dcnn} trained a 2D CNN for aortic thrombus volume assessment from pre-operatively and post-operatively segmentations using rotating and mirroring augmentation.
Postprocessing includes Gaussian filtering and k-means clustering.
In their article Hong et al.\cite{hong2016automatic} trained a DBN using image patches for the detection, segmentation and severity classification of Abdominal Aortic Aneurysm region in CT images.
Liu et al.\cite{liu2017left} used an FCN with twelve \textit{layers} for left atrium segmentation in 3D CT volumes and then refined the segmentation results of the FCN with an active shape model achieving a Dice of 93\%.

CT has also been used for localization of cardiac areas.
In\cite{de20162d} the authors created a method to detect anatomical ROIs (heart, aortic arch, and descending aorta) in 2D image slices from chest CT in order to localize them in 3D.
Every ROI was identified using a combination of three CNNs, each analyzing one orthogonal image plane.
While a single CNN predicted the presence of a specific ROI in the given plane, the combination of their results provided a 3D bounding box around it.
In their article Moradi et al.\cite{moradi2016hybrid} address the problem of detection of vertical position for a given cardiac CT slice.
They divide the body area depicted in chest CT into nine semantic categories each representing an area most relevant to the study of a disease.
Using a set of handcrafted image features together with features derived from a pretrained VGGnet with five \textit{layers}, they build a classification scheme to map a given CT slice to the relevant level.
Each feature group was used to train a separate SVM classifier and predicted labels are then combined in a linear model, also learned from training data.

Deep learning was used with CT from other regions besides the heart.
Zheng et al.\cite{zheng20153d} created a method for 3D detection in volumetric data which they quantitatively evaluated for carotid artery bifurcation detection in CT\@.
An one hidden layer network was used for the initial testing of all voxels to obtain a small number of candidates, followed by a more accurate classification with a deep network.
The learned image features are further combined with Haar wavelet features to increase the detection accuracy.
Montoya et al.\cite{montoya2018deep} trained a 30 layer ResNet to generate 3D cerebral angiograms from contrast-enhanced images using three tissue types (vasculature, bone and soft tissue).
They created the annotations using thresholding and connected components in 3D space, having a combined dataset of 13790 images.

CT has also been used for other tasks.
Zreik et al.\cite{zreik2018deep} created a method to identify patients with coronary artery stenoses from the LV myocardium in rest CT\@.
They used a multi-scale CNN to segment the LV myocardium and then encoded it using an unsupervised convolutional AE\@.
Thereafter, the final classification is done using an SVM classifier based on the extracted and clustered encodings.
Similar work has been done by the same authors in\cite{zreik2016automatic} which they use three CNNs to detect a bounding box around the LV and perform LV voxel classification within the bounding box.
Commandeur et al.\cite{commandeur2018deep} used a combination of two deep networks to quantify epicardial and thoracic apidose tissue in CT from 250 patients with 55 slices per patient on average.
The first network is a six \textit{layer} CNN that detects the slice located within heart limits, and segments the thoracic and epicardial-paracardial masks.
The second network is a five \textit{layer} CNN that detects the pericardium line from the CT scan in cylindrical coordinates.
Then a statistical shape model regularization along with thresholding and median filtering provide the final segmentations.
Gulsun et al.\cite{gulsun2016coronary} created a method for the extraction of blood vessel centerlines in CT\@.
First, optimal paths in a computed flow field are found and then a CNN classifier is used for removing extraneous paths in the detected centerlines.
The method was enhanced using a model-based detection of coronary specific territories and main branches to constrain the search space.

\subsection{Echocardiography}
Echocardiography is an imaging modality that depicts the heart area using ultrasound waves.
Uses of deep learning in echocardiography mainly include LV segmentation and quality score assessment amongst others.

DBNs have been used for LV segmentation in echocardiography.
In\cite{carneiro2012segmentation} the authors created a method that decouples the rigid and nonrigid detections with a DBN that models the appearance of the LV demonstrating that it is more robust than level sets and deformable templates.
Nascimento et al.\cite{nascimento2016multi} used manifold learning that partitions the data into patches that each one proposes a segmentation of the LV\@.
The fusion of the patches was done by a DBN multi-classifier that assigns a weight for each patch.
In that way the method does not rely on a single segmentation and the training process produces robust appearance models without the need of large training sets.
In\cite{chen2016iterative} the authors used a multi-domain regularized FCN and transfer learning.
They compare their method with simpler FCN architectures and a state-of-the-art method demonstrating better results.

Echocardiography has also been used for viewpoint classification.
Madani et al.\cite{madani2018fast} trained a six \textit{layer} CNN to classify between 15 views (12 video and 3 still) of transthoracic echocardiogram images, achieving better results than certified echocardographers.
In\cite{silva2018ejection} the authors created a residual 3D CNN for ejetion fraction classification from transthoracic echocardiogram images.
They used 8715 exams each one with 30 sequential frames of the apical 4 chamber to train and test their method achieving preliminary results.
Gao et al.\cite{gao2017fused} incorporated spatial and temporal information sustained by the video images of the moving heart by fusing two CNNs with seven \textit{layers} each.
The acceleration measurement at each point was calculated using dense optical flow method to represent temporal motion information.
Subsequently, the fusion of the CNNs was done using linear integrations of the vectors of their outputs.
Comparisons were made with previous hand-engineering approaches demonstrating superior results.

Quality score assessment and other tasks were also targeted using echocardiography.
In\cite{abdi2017quality} the authors created a method for reducing operator variability in data acquisition by computing an echo quality score for real-time feedback.
The model consisted of convolutional layers to extract features from the input echo cine and recurrent layers to use the sequential information in the echo cine loop.
Ghesu et al.\cite{ghesu2016marginal} method for object detection and segmentation in the context of volumetric image parsing, is done solving anatomical pose estimation and boundary delineation.
For this task they introduce marginal space deep learning which provides high run-time performance by learning classifiers in clustered, high-probability regions in spaces of gradually increasing dimensionality.
Given the object localization, they propose a combined deep learning active shape model to estimate the non-rigid object boundary.
In their article Perrin et al.\cite{perrin2017application} trained and evaluated AlexNet with 59151 echo frames in a pairwise fashion to classify between five pediatric populations with congenital heart disease.
Moradi et al.\cite{moradi2016cross} created a method based on VGGnet and doc2vec\cite{le2014distributed} to produce semantic descriptors for images which can be used as weakly labeled instances or corrected by medical experts.
Their model was able to identify 91\% of diseases instances and 77\% of disease severity modifiers from Doppler images of cardiac valves.

\subsection{Optical coherence tomography}
Optical Coherence Tomography (OCT) is an intravascular imaging modality that provides cross-sectional images of arteries with high resolution and reproducible quantitative measurements of the coronary geometry in the clinical setting\cite{kubo2013oct}.

In their article Roy et al.\cite{roy2016multiscale} characterized tissue in OCT by learning the multi-scale statistical distribution model of the data with a distribution preserving AE\@.
The learning rule of the network introduces a scale importance parameter associated with error backpropagation.
Compared with three baseline pretrained AEs with cross entropy it achieves better performance in terms of accuracy in plaque/normal pixel detection.
Yong et al.\cite{yong2017linear} created a linear-regression CNN with four \textit{layers} to segment vessel lumen, parameterized in terms of radial distances from the catheter centroid in polar space.
The high accuracy of this method along with its computational efficiency (40.6ms/image) suggest the potential of being used in the real clinical environment.
In\cite{xu2017fibroatheroma} the authors compared the discriminative capability of deep features extracted from each one of AlexNet, GoogleNet, VGG-16 and VGG-19 to identify fibroatheroma.
Data augmentation was applied on a dataset of OCT images for each classification scheme and linear SVM was conducted to classify normal and fibroatheroma images.
Results indicate that VGG-19 is better in identifying images that contain fibroatheroma.
Abdolmanafi et al.\cite{abdolmanafi2017deep} classify tissue in OCT using a pretrained AlexNet as feature extractor and compare the predictions of three classifiers, CNN, RF, and SVM, with the first one achieving the best results.

\subsection{Other imaging modalities}
Intravascular Ultrasound (IVUS) uses ultraminiaturized transducers mounted on modified intracoronary catheters to provide radial anatomic imaging of intracoronary calcification and plaque formation\cite{parrillo2013critical}.
Lekadir et al.\cite{lekadir2017convolutional} used a patched-based four \textit{layer} CNN for characterization of plaque composition in carotid ultrasound images.
Experiments done by the authors showed that the model achieved better pixel-based accuracy than single-scale and multi-scale SVMs.
In\cite{tajbakhsh2017automatic} the authors automated the entire process of carotid intima media thickness video interpretation.
They trained a two \textit{layer} CNN with two outputs for frame selection and a two \textit{layer} CNN with three outputs for ROI localization and intima-media thickness measurements.
This model performs much better than a previous handcrafted method by the same authors, which they justify in CNN's capability to learn the appearance of QRS and ROI instead of relying on a thresholded R-peak amplitude and curvature.
Tom et al.\cite{tom2018simulating} created a Generative Adversarial Network (GAN) based method for fast simulation of realistic IVUS\@.
Stage 0 simulation was performed using pseudo B-mode IVUS simulator and yielded speckle mapping of a digitally defined phantom.
Stage I refined the mappings to preserve tissue specific speckle intensities using a GAN with four residual blocks and Stage II GAN generated high resolution images with patho-realistic speckle profiles.

Other cardiology related applications with deep learning use modalities such as mammograms, X-Ray, percutaneous transluminal angioplasty, biopsy images and myocardial perfusion imaging.
In their article Wang et al.\cite{wang2017detecting} apply a pixelwise, patch-based procedure for breast arterial calcification detection in mammograms using a ten \textit{layer} CNN and morphologic operation for post-processing.
The authors used 840 images and their experiments resulted in a model that achieved a coefficient of determination of 96.2\%.
Liu et al.\cite{liu2017coronary} trained CNNs using 1768 X-Ray images with corresponding diagnostic reports.
The average diagnostic accuracies of the models reached to a maximum of 0.89 as the depth increased reaching eight layers; after that the increase in accuracy was limited.
In\cite{pavoni2017image} the authors created a method to denoise low dose percutaneous transluminal coronary angioplasty images.
They tested mean squared error and structural similarity based loss functions on two patch-based CNNs with four \textit{layers} and compared them for different types and levels of noise.
Nirschl et al.\cite{nirschl2018deep} used endomyocardial biopsy images from 209 patients to train and test a patch-based six \textit{layer} CNN to identify heart failure.
Rotational data augmentation was also used while the outputs of the CNN on each patch were averaged to obtain the image-level probability.
This model demonstrated better results than AlexNet, Inception and ResNet50.
In\cite{betancur2018deep} the authors trained a three \textit{layer} CNN for the prediction of obstructive CAD stress myocardial perfusion from 1638 patients and compared it with total perfusion deficit.
The model computes a probability per vessel during training, while in testing the maximum probabilities per artery are used per patient score.
Results show that this method outperforms the total perfusion deficit in the per vessel and per patient prediction tasks.

\subsection{Overall view on deep learning using CT, echocardiography, OCT and other imaging modalities}
Papers using these imaging modalities were highly variable in terms of the research question they were trying to solve and highly inconsistent in respect with the use of metrics for the results they reported.
These imaging modalities also lack behind in having publicly available database thus limiting opportunities for new architectures to be tested by groups that do not have an immediate clinical partner.
On the other hand there is relatively high uniformity regarding the use of architectures with CNN most widely used, especially the pretrained top performing architectures from the ImageNet competition (AlexNet, VGG, GoogleNet, ResNet).

\section{Discussion and future directions}
\label{sec:discussion}

\begin{table*}[!t]
	\caption{Reviews of deep learning applications in cardiology}
	\label{table:reviews}
	\centering
	\begin{tabularx}{\textwidth}{l l}
		\toprule
		\thead{Reference}                                  & \thead{Application/Notes}                                                                                                 \\
		\midrule
		Mayer 2015\cite{mayer2015big}                      & Big data in cardiology changes how insights are discovered                                                                \\
		Austin 2016\cite{austin2016application}            & overview of big data, its benefits, potential pitfalls and future impact in cardiology                                    \\
		Greenspan 2016\cite{greenspan2016guest}            & lesion detection, segmentation and shape modeling                                                                         \\
		Miotto 2017\cite{miotto2017deep}                   & imaging, EHR, genome and wearable data and needs for increasing interpretability                                          \\
		Krittanawong 2017\cite{krittanawong2017rise}       & studies on image recognition technology which predict better than physicians                                              \\
		Litjens 2017\cite{litjens2017survey}               & image classification, object detection, segmentation and registration                                                     \\
		Qayyum 2017\cite{qayyum2017medical}                & CNN-based methods in image segmentation, classification, diagnosis and image retrieval                                    \\
		Hengling 2017\cite{henglin2017machine}             & impact that machine learning will have on the future of cardiovascular imaging                                            \\
		Blair 2017\cite{blair2017advanced}                 & advances in neuroimaging with MRI on small vessel disease                                                                 \\
		Slomka 2017\cite{slomka2017cardiac}                & nuclear cardiology, CT angiography, Echocardiography, MRI                                                                 \\
		Carneiro 2017\cite{carneiro2017review}             & mammography, cardiovascular and microscopy imaging                                                                        \\
		Johnson 2018\cite{johnson2018artificial}           & AI in cardiology describing predictive modeling concepts, common algorithms and use of deep learning                      \\
		Jiang 2017\cite{jiang2017artificial}               & AI applications in stroke detection, diagnosis, treatment, outcome prediction and prognosis evaluation                    \\
		Lee 2017\cite{lee2017deepb}                        & AI in stroke imaging focused in technical principles and clinical applications                                            \\
		Loh 2017\cite{loh2017deep}                         & heart disease diagnosis and management within the context of rural healthcare                                             \\
		Krittanawong 2017\cite{krittanawong2017artificial} & cardiovascular clinical care and role in facilitating precision cardiovascular medicine                                   \\
		Gomez 2018\cite{gomez2018new}                      & recent advances in automation and quantitative analysis in nuclear cardiology                                             \\
		Shameer 2018\cite{shameer2018machine}              & promises and limitations of implementing machine learning in cardiovascular medicine                                      \\
		Shrestha 2018\cite{shrestha2018machine}            & machine learning applications in nuclear cardiology                                                                       \\
		Kikuchi 2018\cite{kikuchi2018future}               & application of AI in nuclear cardiology and the problem of limited number of data                                         \\
		Awan 2018\cite{awan2018machine}                    & machine learning applications in heart failure diagnosis, classification, readmission prediction and medication adherence \\
		Faust 2018\cite{faust2018deep}                     & deep learning application in physiological data including ECG                                                             \\
		\bottomrule
	\end{tabularx}
\end{table*}

It is evident from the literature that deep learning methods will replace rule-based expert systems and traditional machine learning based on feature engineering.
In\cite{awan2018machine} the authors argue that deep learning is better in visualizing complex patterns hidden in high dimensional medical data.
Krittanawong et al.\cite{krittanawong2017artificial} argue that the increasing availability of automated real-time AI tools in EHRs will reduce the need for scoring systems such as the Framingham risk.
In\cite{krittanawong2017rise} the authors argue that AI predictive analytics and personalized clinical support for medical risk identification are superior to human cognitive capacities.
Moreover, AI may facilitate communication between physicians and patients by decreasing processing times and therefore increasing the quality of patient care.
Loh et al.\cite{loh2017deep} argue that deep learning and mobile technologies would expedite the proliferation of healthcare services to those in impoverished regions which in turn leads to further decline of disease rates.
Mayer et al.\cite{mayer2015big} state that big data promises to change cardiology through an increase in the data gathered but its impact goes beyond improving existing methods such as changing how insights are discovered.

Deep learning requires large training datasets to achieve high quality results\cite{krizhevsky2012imagenet}.
This is especially difficult with medical data, considering that the labeling procedure of medical data is costly because it requires manual labor from medical experts.
Moreover most of medical data belong to the normal cases instead to abnormal, making them highly unbalanced.
Other challenges of applying deep learning in medicine that previous literature has identified are data standardization/availability/dimensionality/volume/quality issues, difficulty in acquiring the corresponding annotations and noise in annotations\cite{litjens2017survey, greenspan2016guest, miotto2017deep, slomka2017cardiac}.
More specifically, in\cite{blair2017advanced} the authors note that deep learning applications on small vessel disease have been developed using only a few representative datasets and they need to be evaluated in large multi-center datasets.
Kikuchi et al.\cite{kikuchi2018future} mention that compared with CT and MRI, nuclear cardiology imaging modalities have limited number of images per patient and only specific number of organs are depicted.
Liebeskind\cite{liebeskind2018artificial} states that machine learning methods are tested on selective and homogeneous clinical data but generalizability would occur using heterogeneous and complex data.
The example of ischemic stroke is mentioned as an example of an heterogeneous and complex disease where the occlusion of middle cerebral artery can lead to divergent imaging patterns.
In\cite{gomez2018new} the authors conclude that additional data that validate these applications in multi-center uncontrolled clinical settings are required before implementation in routine clinical use.
The impact of these tools on decision-making, down-stream utilization of resources, cost, and value-based practice also needs to be investigated.
Moreover the present literature demonstrated that there is an unbalanced distribution of publicly available datasets among different imaging modalities in cardiology (e.g.\ no public dataset available for OCT in contrast with MRI).

Previous literature states that problems related to data can be solved using data augmentation, open collaboration between research organizations and increase in funding.
Hengling et al.\cite{henglin2017machine} argue that substantial investments will be required to create high quality annotated databases which are essential for the success of supervised deep learning methods.
In\cite{austin2016application} the authors argue that the continued success of this field depends on sustained technological advancements in information technology and computer architecture as well as collaboration and open exchange of data between physicians and other stakeholders.
Lee et al.\cite{lee2017deepb} conclude that international cooperation is required for constructing a high quality multimodal big dataset for stroke imaging.
Another solution to better exploit big medical data in cardiology is to apply unsupervised learning methods, which do not require annotations.
The present review demonstrated that unsupervised learning is not thoroughly used since the majority of the methods in all modalities are supervised.

Regarding the problem of lack of interpretability as it is indicated by Hinton\cite{hinton2018deep} it is generally infeasible to interpret nonlinear features of deep networks because their meaning depends on complex interactions with uninterpreted features from other layers.
Additionally these models are stochastic, meaning that every time a network fits the same data but with different initial weights different features are learned.
More specifically, in an extensive review\cite{betancur2018deep} of whether the problem of LV/RV segmentation is solved the authors state that although the classification aspect of the problem achieves near perfect results the use of a `diagnostic black box' can not be integrated in the clinical practice.
Miotto et al.\cite{miotto2017deep} mention interpretability as one of the main challenges facing the clinical application of deep learning to healthcare.
In\cite{lee2017deepb} the authors note that the black-box nature of AI methods like deep learning is against the concept of evidence-based medicine and it raises legal and ethical issues in using them in clinical practice.
This lack of interpretability is the main reason that medical experts resist using these models and there are also legal restrictions regarding the medical use of the non-interpretable applications\cite{slomka2017cardiac}.
On the other hand, any model can be placed in a `human-machine decision effort' axis\cite{beam2018big} including statistical ones that medical experts rely on for everyday clinical decision making.
For example, human decisions such as choosing which variables to include in the model, the relationship of dependent and independent variables and variable transformations, move the algorithm to the human decision axis, thus making it more interpretable but in the same time more error-prone.

Regarding the solution to the interpretability problem when new methods are necessary researchers should prefer making simpler deep learning methods (end-to-end and non-ensembles) to increase their clinical applicability, even if that means reduced reported accuracy.
There are also arguments against creating new methods but instead focus on validating the existing ones.
In\cite{damen2016prediction} the authors conclude that there is an excess of models predicting incident CVD in the general population.
Most of the models usefulness is unclear due to errors in methodology and lack of external validation studies.
Instead of developing new CVD risk prediction models future research should focus on validating and comparing existing models and investigate whether they can be improved.

A popular method used for interpretable models is attention networks\cite{bahdanau2014neural}.
Attention networks are inspired by the ability of human vision to focus in a certain point with high resolution while perceiving the surroundings with low resolution and then adjusting the focal point.
They have been used by a number of publications in cardiology in medical history prediction\cite{kim2017highrisk}, ECG beat classification\cite{schwab2017beat} and CVD prediction using fundus\cite{poplin2017predicting}.
Another simpler tool for interpretability is saliency maps\cite{simonyan2013deep} that uses the gradient of the output with respect to the input which intuitively shows the regions that most contribute toward the output.

Besides solving the data and interpretability problems, researchers in cardiology could utilize the already established deep learning architectures that have not been widely applied in cardiology such as capsule networks.
Capsule networks\cite{sabour2017dynamic} are deep neural networks that require less training data than CNNs and its layers capture the `pose' of features thus making their inner-workings more interpretable and closer to the human way of perception.
However an important constraint they currently have which limits them from achieving wider use, is the high computational cost compared to CNNs due to the `routing by agreement' algorithm.
Amongst their recent uses in medicine include brain tumor classification\cite{afshar2018brain} and breast cancer classification\cite{iesmantas2018convolutional}.
Capsule networks have not been used in cardiology data yet.

Another underused deep learning architecture in cardiology is GANs\cite{goodfellow2014generative} that consist of a generator that creates fake images from noise and a discriminator that is responsible of differentiating between fake images from the generator and real images.
Both networks try to optimize a loss in a zero-sum game resulting in a generator that produces realistic images.
GANs have only been used for simulating patho-realistic IVUS images\cite{tom2018simulating} and the cardiology field has a lot to gain from using this kind of models, especially in the absense of high quality annotated data.

Researchers could also utilize CRFs which are graphical models that capture context-aware information and are able to incorporate higher order statistics, which traditional deep learning methods are unable to do.
CRFs have been jointly trained with CNNs and have been used in depth estimation in endoscopy\cite{mahmood2018deep} and liver segmentation in CT\cite{christ2016automatic}.
There are also cardiology applications that used CRFs with deep learning as a segmentation refinement step in fundus photography\cite{zhou2017improving, fu2016retinal}, and in LV/RV\cite{bai2017semi}.
Multimodal deep learning\cite{ngiam2011multimodal} can also be used to improve diagnostic outcomes e.g.\ the possibility of combining fMRI and ECG data.
Dedicated databases must be created in order to increase research in this area since according to the current review there are only three cardiology databases with multimodal data.
In addition to the previous databases MIMIC-III has also been used for multimodal deep learning by \cite{purushotham2018benchmarking} for predicting in-hospital, short/long-term mortality and ICD-9 code predictions.

\section*{Conclusions}
With each technological advance, cardiology and medicine in general is becoming human independent and closer to an automated AI driven field.
AI will not only reach the point where it uses real-time physical scans to detect diseases, but it will also interpret ambiguous conditions, precisely phenotype complex diseases and take medical decisions.
However, a complete theoretical understanding of deep learning is not yet available and a critical understanding of the strengths and limitations of its inner workings is vital for the field to gain its place in everyday clinical use.
Successful application of AI in the medical field relies on achieving interpretable models and big datasets.

\clearpage{}
\bibliographystyle{IEEEtran}
\bibliography{ms.bib}

\end{document}